\documentclass{article}
\usepackage{ijcai22}
\usepackage{times}

\usepackage{soul}
\usepackage{url}
\usepackage[utf8]{inputenc}
\usepackage[small]{caption}
\usepackage[list=true]{subcaption}
\usepackage{graphicx}
\usepackage{amsmath}
\usepackage{amssymb}
\usepackage{booktabs}
\usepackage{tabu}
\usepackage{wrapfig}
\usepackage{arydshln}
\usepackage[titletoc]{appendix}
\usepackage{bibunits}
\usepackage[colorlinks=true]{hyperref}
\usepackage[textsize=tiny]{todonotes}
\usepackage{siunitx}
\usepackage{multirow}
\usepackage[normalem]{ulem}
\usepackage{bbm}
\usepackage{scrextend}
\deffootnote[2em]{2em}{1em}{\textsuperscript{\thefootnotemark}\,}

\newcommand\freefootnote[1]{
  \let\thefootnote\relax
  \footnotetext{#1}
  \let\thefootnote\svthefootnote
}

\title{Uncertainty-aware Evaluation of Time-Series Classification for Online Handwriting Recognition with Domain Shift}

\author{
Andreas Klaß$^{1,2,}$\footnote{Equal contribution}\and
Sven M. Lorenz$^{1,2,*}$\and
Martin W. Lauer-Schmaltz$^{4}$\and
David Rügamer$^{2,3}$\and \\
Bernd Bischl$^{2}$\and
Christopher Mutschler$^{1}$\And
Felix Ott$^{1,2}$\\
\affiliations
$^1$Fraunhofer IIS, Fraunhofer Institute for Integrated Circuits IIS\\
$^2$LMU Munich, Munich, Germany\\
$^3$RWTH Aachen, Aachen, Germany\\
$^4$Technical University of Denmark\\
\emails
\{a.klass, sven.lorenz\}@campus.lmu.de,
\{david.ruegamer, bernd.bischl\}@stat.uni-muenchen.de,\\
\{christopher.mutschler, felix.ott\}@iis.fraunhofer.de,
\{mwola\}@dtu.dk
}

\begin{document}

\maketitle
\begin{abstract}
For many applications, analyzing the uncertainty of a machine learning model is indispensable. While research of uncertainty quantification (UQ) techniques is very advanced for computer vision applications, UQ methods for spatio-temporal data are less studied. In this paper, we focus on models for online handwriting recognition, one particular type of spatio-temporal data. The data is observed from a sensor-enhanced pen with the goal to classify written characters. We conduct a broad evaluation of aleatoric (data) and epistemic (model) UQ based on two prominent techniques for Bayesian inference, Stochastic Weight Averaging-Gaussian (SWAG) and Deep Ensembles. Next to a better understanding of the model, UQ techniques can detect out-of-distribution data and domain shifts when combining right-handed and left-handed writers (an underrepresented group).\freefootnote{"Copyright © 2022 for this paper by its authors. Use permitted under Creative Commons License Attribution 4.0 International (CC BY 4.0)."}
\end{abstract}
\section{Introduction}
\label{chap_introduction}

Traditional machine learning (ML) algorithms assume training and test datasets to be \textit{independently and identically distributed} \cite{sun_feng,schoelkopf_locatello}. For many real-world applications, data often changes over time and space, and hence, training and test data originate from different distributions. This can cause ML models to fail due to a \emph{domain shift} between training and test data \cite{sun_feng}. Transfer learning \cite{pan_yang,shao_zhu} and domain adaptation \cite{long_wang,saenko} techniques can compensate for this domain shift. A first step in adapting for this domain shift is its detection, e.g., by having reliable uncertainty estimates of the model predictions \cite{li2022uncertainty}. Thus, to estimate the uncertainty of the model, a robust uncertainty quantification (UQ) technique is required that runs in real-time.

\paragraph{Approximate Bayesian Inference Techniques.} In the field of deep learning (DL), UQ has lately seen a steep increase in interest. Recently, many promising methods have been proposed such as Variational Online Gauss-Newton (VOGN) \cite{khan2018fast}, Stochastic Weight Averaging-Gaussian (SWAG) \cite{maddox2019simple}, Bayes by Backpropagation (BBB) \cite{blundell}, and Laplace Approximation \cite{daxberger_kristiadi}. Another widely used technique are Deep Ensembles \cite{lakshminarayanan2017simple}, which often yield well-calibrated models while being relatively easy to implement.

\paragraph{Decomposing Uncertainty.} Several ways for estimating and decomposing uncertainty have been proposed. A common distinction is made between \textit{aleatoric} uncertainty, which refers to the variability in the data, and \textit{epistemic} uncertainty, which is the model's uncertainty caused by a lack of knowledge \cite{huellermeier_waegeman}. Building on \cite{kendall2017uncertainties}, \cite{kwon2018uncertainty} argue that neural networks (NNs) for classification are basically generalized linear models with error structure of multinomial and composite link functions. Hence, to acknowledge that the variance of a multinomial outcome is a function of the mean outcome, they propose to directly compute the variability in the softmax outputs. Another method to dissect total predictive uncertainty has been put forward by \cite{smith2018understanding} and similarly by \cite{depeweg2018decomposition} who propose to extract epistemic and aleatoric uncertainties from the predictive distribution of a Bayesian NN by calculating the entropy and mutual information. For an extensive survey of related approaches, see \cite{gawlikowski2022survey}.

\paragraph{UQ for OnHW.} UQ techniques have been broadly evaluated in computer vision applications such as image classification \cite{kendall2017uncertainties}, i.e., optical character recognition (OCR), but methods have rarely been evaluated on spatio-temporal datasets \cite{cai_chen}. OCR is concerned with offline handwriting recognition from images. In contrast, online handwriting (OnHW) recognition works on different types of spatio-temporal signals and can make use of temporal information such as writing speed and direction \cite{plamondon}. While many recording systems make use of a stylus pen together with a touch screen surface, sensor-enhanced pens, e.g., \cite{ott,ott_ijcai,ott_wacv,ott_ijdar}, based on inertial measurement units (IMUs) enable new applications. These pens stream data from accelerometer, gyroscope, magnetometer and force sensors in real-time represented as spatio-temporal multivariate time-series (MTS). The advantage of exploiting this temporal information is the ability to better distinguish between similarly shaped letters from dynamic information (number of strokes etc.). Spatio-temporal data can further help to identify certain characteristics in the data. \cite{ott_mm}, e.g., showed the domain shift between right-handed and left-handed writers by analyzing feature embeddings of their model for OnHW data.

\paragraph{Contribution.} In this paper we evaluate the uncertainty of OnHW model predictions with SWAG \cite{maddox2019simple} and Deep Ensembles \cite{lakshminarayanan2017simple} for spatio-temporal reasoning, assessment of out-of-distribution detection, and pattern and failure recognition. We use uncertainty decompositions based on the method by \cite{kwon2018uncertainty} and \cite{smith2018understanding} to evaluate the UQ techniques. Our claims are further supported by utilizing confidence and accuracy metrics to estimate the expected calibration error (ECE) \cite{guo2017calibration}. For an OnHW task with domain shift between right- and left-handed writers, we evaluate uppercase, lowercase and combined character classification tasks. Our source code will be available upon publication.\footnote{Code and datasets: \href{https://www.iis.fraunhofer.de/de/ff/lv/dataanalytics/anwproj/schreibtrainer/onhw-dataset.html}{www.iis.fraunhofer.de/de/ff/lv/dataanalytics/ anwproj/schreibtrainer/onhw-dataset.html}}

The remainder of the paper is organized as follows. Section~\ref{chap_related_work} discusses related work. In Section~\ref{chap_method}, we describe the background of Bayesian modeling and approximate inference. The experimental setup is described in Section~\ref{chap_experiments}, and results are discussed in Section~\ref{chap_evaluation}.

\section{Related Work}
\label{chap_related_work}

We first present related work of UQ with focus on spatio-temporal reasoning in Section~\ref{section_rw_uq}. Section~\ref{section_rw_onhw} summarizes state-of-the-art results for OnHW recognition.

\subsection{UQ for Spatio-Temporal Reasoning}
\label{section_rw_uq}

\cite{wu_gao_xiong} analyzed Bayesian and frequentist UQ methods for spatio-temporal forecasting on network traffic, epidemics and air quality datasets. Their evaluation shows that Bayesian methods are typically more robust in mean prediction, while confidence levels from frequentist methods provide better coverage over data variations (i.e., out-of-distribution data). Furthermore, traditional learning schemes lack knowledge about uncertainty. STUaNet~\cite{zhou_wang_xie} addresses this issue for spatio-temporal human mobility forecasting by injecting controllable uncertainty. This allows insights to both, UQ and weak supervised learning. \cite{gomez_guan_tripathy} focused on the spatio-temporal uncertainty of urban prediction (where and when a piece of land becomes urban). \cite{li2022uncertainty} argue that the feature statistics such as mean and standard deviation (the domain characteristics of the training data), can be manipulated to improve the generalizability of DL models by modeling the uncertainty of domain shifts with feature statistics during training (that follow a multivariate Gaussian distribution). In the context of domain adaptation, \cite{cai_chen} adressed the extraction of domain-invariant representations for MTS classification.

\subsection{Online Handwriting Recognition}
\label{section_rw_onhw}

\cite{ott} initially proposed the \textit{OnHW-chars} dataset and evaluated machine and DL techniques for the OnHW MTS classification task. The dataset contains right-handed and left-handed writers with a domain shift between both groups of writers (i.e., domains). \cite{ott_mm} showed that transfer learning from small adaptation datasets results in poor model performances. Hence, their domain adaptation approach transforms features from left-handed writers into the domain of features from right-handed writers by optimal transport techniques. A reliable UQ method could identify out-of-distribution samples and only apply the transformation on samples for which the model has a high uncertainty. \cite{ott_ijcai} combined offline and online handwriting recognition with a cross-modal representation learning technique by increasing the dataset size by using generative models. A robust uncertainty estimation technique could select samples with high model uncertainty.
\section{Methodological Background}
\label{chap_method}

In the following we describe Bayesian model averaging in Section~\ref{bayesian_model_avg} and the two employed Bayesian UQ methods in Section~\ref{chap_exp_bnn}. The decomposition of total predictive uncertainty into aleatoric and epistemic uncertainty is discussed in Section~\ref{section_uq_al_ep}.

\subsection{Bayesian Model Averaging}
\label{bayesian_model_avg}

Bayesian approaches in DL naturally represent uncertainty by placing a distribution over model parameters and then marginalizing these parameters to form a predictive distribution (\textit{Bayesian model averaging}) \cite{hoeting1999bayesian}. Let $p(\theta|D)$ be the posterior distribution over model parameters $\theta$, i.e., real-valued weights in the NN, given training dataset $D$, and let $p(y^*|x^*, \theta)$ denote the probability distribution over model outputs $y^*$ (predicted classes), given sample $x^*$, and model weights $\theta$. For the OnHW classification task, the sample $x^*$ is an MTS $\mathbf{U} = \{\mathbf{u}_1,\ldots,\mathbf{u}_q\} \in \mathbb{R}^{q \times l}$, an ordered sequence of $l = 13$ streams with $\mathbf{u}_i = (u_{i,1},\ldots, u_{i,l}), i\in \{1, \ldots, q\}$, where $q = 64$ is the length of the MTS. The training set $D$ is a subset of the array $\mathcal{U} = \{\mathbf{U}_1,\ldots,\mathbf{U}_{n_U}\} \in \mathbb{R}^{n_U \times q \times l}$, where $n_U$ is the number of time-series. The aim is to predict an unknown class label $y^* \in \mathcal{Y}$ with $K$ classes (i.e., character labels) for a given MTS. The predictive distribution of the target variable is then given by
\begin{equation}\label{pred_post}
    p(y^*|x^*, D) = \int p(y^*|x^*, \theta) p(\theta | D) d \theta.
\end{equation}
In practice, we can approximate this integral by drawing $S$ Monte Carlo samples from the posterior distribution:
\begin{equation}
    p(y^*|x^*, D) \approx \frac{1}{S} \sum^{S}_{s=1} p(y^*|x^*, \theta_s) \hspace{0.2cm} , \hspace{0.2cm} \theta_s \sim p(\theta | D).
\end{equation}
The predicted probability of an outcome is thus a weighted average over its probabilities with the weights being determined by $p(\theta |D)$.

\subsection{Approximate Bayesian Inference}
\label{chap_exp_bnn}

In order to apply Bayesian inference to an NN, we need to compute the posterior $p(\theta|D)$ of the NN weights. As the computation of the posterior is usually intractable, a (local) approximation is often used. This can be addressed by SWAG and Deep Ensembles with the latter abstaining from explicitly modeling $p(\theta|D)$ -- nevertheless, this method can be considered to be in the field of approximate Bayesian inference.

\paragraph{Stochastic Weight Averaging-Gaussian (SWAG).} SWAG \cite{maddox2019simple} is a Bayesian inference technique for DL that builds on Stochastic Weight Averaging (SWA) \cite{izmailov2018averaging}. SWA computes an average of stochastic gradient decent (SGD) iterates to obtain information about the geometry of $p(\theta|D)$ from its trajectory. This posterior is then approximated by a Gaussian with simplified covariance structure and reduced dimensionality.

\paragraph{Deep Ensembles.} Deep Ensembles are a committee of individual NNs initialized with a different seed \cite{lakshminarayanan2017simple}. The initialization serves as the only source of stochasticity in the model parameters which are otherwise not random; Deep Ensembles can optionally be coupled with a differently shuffled data loader. In contrast to SWAG, results are obtained by averaging the predictions of $M$ independently trained networks instead of explicitly modeling a posterior and sampling from it. \cite{ovadia2019can} point out that even an ensemble size of $M = 5$ performs well, strengthening its reputation as a ``gold standard'' for accurate and well-calibrated predictive distributions.

\subsection{Uncertainty Decomposition}
\label{section_uq_al_ep}

In the literature two sources of uncertainty are commonly considered \cite{huellermeier_waegeman}: (1) \textit{Aleatoric} uncertainty represents stochasticity inherent in the data. For the OnHW application this can be sensor noise induced by the ballpoint pen on the paper or by shaky hands of the writer. In particular, even with infinitely many data points, there will always be some variation in the data. (2) \textit{Epistemic} uncertainty is the model uncertainty, which, in theory, can be reduced to zero for an increasing amount of observations. Various approaches of measuring uncertainty exist in the literature. We consider two approaches, both providing justified and mutually complementing insights into our trained models and data situation: uncertainty decomposition based on the softmax output variability \cite{kwon2018uncertainty} in Section~\ref{sub_sectin_ud_knwon} and based on information theory in Section~\ref{section_uq_inform}.

\subsubsection{Uncertainty Decomposition based on [Kwon et al.]} 
\label{sub_sectin_ud_knwon}

The definition proposed by \cite{kwon2018uncertainty} is based on considerations by \cite{kendall2017uncertainties} and presents a novel way to estimate predictive uncertainty by breaking it down into
\begin{equation}
\label{kwon}
    \underbrace{\frac{1}{T} \sum_{t=1}^{T} \text{diag}(\hat{c}_t) - \hat{c}_t\hat{c}_t^\top}_{\text{aleatoric uncertainty}} + \underbrace{\frac{1}{T} \sum_{t=1}^{T} (\hat{c}_t - \bar{c})(\hat{c}_t - \bar{c})^\top}_{\text{epistemic uncertainty}},
\end{equation}
with $\hat{c}_t = (\hat{c}_{t,1}, \ldots ,\hat{c}_{t,K}) \in [0,1]^K$ being the softmax output of the NN based on one forward pass (out of $T$ stochastic forward passes), $\sum_{i=1}^{K} \hat{c}_{t,i} = 1$, and $\bar{c} = \frac{1}{T} \sum_{t=1}^{T} \hat{c}_t$.

\paragraph{Interpretation.} Equation~\ref{kwon} yields two $K \times K$ matrices with different interpretations. For the \textit{aleatoric} part, diagonal values are in $\{x-x^2 \mid x \in [0,1]\}$, with the maximum uncertainty for $x = 0.5$. If the model is ``unsure'', meaning that the model neither displays confidence that a prediction corresponds to a certain class nor displays confidence that it is not, we expect high aleatoric uncertainty. The off-diagonal elements consist of values in $\{-x \cdot y \mid x,y \in [0,1]\}$, which yields values on the interval $[-0.25,0]$. Lower values correspond to higher data uncertainty. For the \textit{epistemic} part, the diagonal contains the squared difference to the mean softmax outputs (over $T$ samples). The off-diagonal has positive values when the softmax values coincide and negative values if the softmax values display an inverse relationship.

\subsubsection{Uncertainty Decomposition based on Information Theory}
\label{section_uq_inform}

Another way to decompose predictive uncertainty into an aleatoric and epistemic part is by following \cite{depeweg2018decomposition} and similarly \cite{smith2018understanding}. Based on principles from information theory, the Shannon entropy $H(p)=-\sum_{i=1}^{K}p_i log_2(p_i)$ is utilized as a common measure of ``informedness'' of a single probability distribution $p$ with $K$ outcomes/classes and the associated probabilities for each $i$-th class $p_i$; taking the logarithm to base 2 yields values measured in \textit{bits}. The total predictive uncertainty (TU) of the predictive distribution $p(y^*|x, D)$ can then be quantified by 
\begin{equation}
    TU = H\big(p(y^*|x^*, D)\big) \approx H\Big({\frac{1}{S} \sum^{S}_{s=1} p(y^*|x^*, \theta_s)}\Big).
\end{equation}
Effectively, this is the entropy of the averaged categorical predictions, and it includes the two sources of uncertainty we are interested in.

\paragraph{Aleatoric Uncertainty (AU), Entropy.} We can express aleatoric uncertainty as the expectation over the entropies of $S$ sampled conditional predictive distributions with fixed weights, i.e.,
\begin{equation}
AU \approx \frac{1}{S} \sum^{S}_{s=1} H(p(y^*|x^*, \theta_s)).
\end{equation}

\paragraph{Epistemic Uncertainty (EU), Mutual Information.} Finally, epistemic uncertainty emerges as the difference of total and aleatoric uncertainty $EU  = TU - AU$, and is equivalent to the mutual information (MI):
\begin{equation}
    \resizebox{.91\linewidth}{!}{$
    \displaystyle
    EU = H\Big({\frac{1}{S} \sum^{S}_{s=1} p(y^*|x^*, \theta_s)}\Big) - \frac{1}{S} \sum^{S}_{s=1} H({p(y^*|x^*, \theta_s)}).
    $}
\end{equation}
Intuitively, epistemic uncertainty stands for the information gain about the model parameters that would be obtained when observing the true outcome. MI is always non-negative, zero in case of perfect independence of $y^*$ and $\theta$, and positive when model uncertainty is present at prediction time.

\section{Experiments}
\label{chap_experiments}

In our order to evaluate the efficacy of UQ methods for spatio-temporal handwriting datasets, we use the OnHW dataset (Section~\ref{chap_exp_data}) and fit different network architectures (Section~\ref{chap_exp_cnns}). Our evaluation approach is given in Section~\ref{chap_exp_metrics}. For architecture and training details and SWAG parameters, see Appendix~\ref{sec_app_parameters}. For Deep Ensembles, we choose $M = 10$ (for a study on number of base learners in Deep Ensembles vs.~SWAG performance, see \cite{maddox2019simple}).

\subsection{Online Handwriting Recognition}
\label{chap_exp_data}

The \textit{OnHW-chars}~\cite{ott} dataset consists of recordings from a sensor-enhanced ballpoint pen providing 14 sensor measurements: two accelerometers (3 axes each), one gyroscope (3 axes), one magnetometer (3 axes), a force sensor (with which the pen tip touches the surface), and the time steps. 119 right-handed and nine left-handed writers participated in the data collection. Each person was instructed to write the English alphabet on plain paper six times. This results in 31,275 right-handed and 2,270 left-handed samples. The task is to either classify lowercase letters (26 classes), uppercase letters (26 classes) or combined letters from all 52 classes. For model evaluation, five cross-validation sets are provided by \cite{ott} for both writer-dependent (WD) and writer-independent (WI) MTS classification tasks.

\subsection{Neural Network Architectures}
\label{chap_exp_cnns}

We use a modified CNN from \cite{ott,ott_ijdar} for feature extraction and combine it with one unit for spatio-temporal classification to extract important temporal features. This unit is added before the last dense layer. We compare a standard long short-term memory (LSTM) cell with 100 neurons, a bidirectional LSTM (BiLSTM) cell with 100 neurons, and a temporal convolutional network (TCN) with 120 neurons. The last dense layer contains 26 neurons for the lowercase and uppercase tasks, or 52 neurons for the combined task. We interpolate the time-series to 64 time steps without sensor normalization.

\subsection{Evaluation Metrics}
\label{chap_exp_metrics}

\paragraph{Confidence Calibration.} Calibration can be understood as the degree of reliability of a model. According to \cite{gawlikowski2022survey}, a predictor is well-calibrated if the derived predictive confidence represents a good approximation of the actual probability of correctness -- meaning that 20\%
of all predictions with a predictive confidence of 80\% should actually be false. Calibration is thus a notion of uncertainty, measuring the discrepancy between the model's forecasts and (empirical) long-run frequencies \cite{lakshminarayanan2017simple}. Using the definitions of confidence and accuracy \cite{guo2017calibration}, we can make statements about over- and under-confidence of the model. We have
\begin{equation}
\label{eq:conf}
    \text{confidence}(b_e) = \frac{1}{|b_e|}\sum_{s\in b_e}\hat{c}_s
\end{equation}
and 
\begin{equation}
\label{eq:acc}
    \text{accuracy}(b_e)=\frac{1}{|b_e|}\sum_{s\in b_e}\mathbbm{1}(\hat{y}_s=y_s),
\end{equation}
with $b_e$ denoting the set of indices of sampled softmax outputs falling into the interval $(l_e, u_e]$. Commonly, the softmax output range is divided into ten bins (interval sizes of 0.1). We can now make statements whether our model is under-confident $\big(\text{accuracy}(b_e) > \text{confidence}(b_e)\big)$ or over-confident $\big(\text{accuracy}(b_e) < \text{confidence}(b_e)\big)$. It has been shown that softmax outputs of deep NNs are in general not well calibrated and are often either over- or under-confident \cite{guo2017calibration}. Ideally, $\text{accuracy}(b_e) \approx \text{confidence}(b_e)$, allowing the user to interpret softmax outputs as probabilities and thereby quantify the prediction uncertainty.

\paragraph{Expected Calibration Error (ECE).} The ECE summarizes how far away the confidence is from the actual (empirical) accuracy \cite{guo2017calibration}. It can be defined as
\begin{equation}
    \resizebox{.91\linewidth}{!}{$
    \displaystyle
    \text{ECE}(b_e)=\sum_{e=1}^{E}\frac{|b_e|}{n} |\text{accuracy}(b_e)-\text{confidence}(b_e)|,
    $}
\end{equation}
with $n$ being the number of predicted softmax outputs, and $E$ being the number of bins. Note that this metric does not give any information about over- or under-confidence -- only how far away the expected accuracy is from the confidence. Ideally, the ECE is 0.

\paragraph{Reliability Diagrams.}\label{reliability_plots}
We visualize Equations~\ref{eq:conf} and \ref{eq:acc} as reliability diagrams \cite{Degroot1983TheCA} for selected models. Generally, a model is over-confident if the black bars (displaying the accuracy for one bin) are below the dashed bisectors. Consequently, if the black bars are above the bisectors, the model is under-confident. We additionally plot the histogram \cite{hollance2020rel} of the softmax outputs to get an overview of the distribution.
\section{Experimental Results}
\label{chap_evaluation}

\begin{table}[t!]
\centering
\setlength{\tabcolsep}{2.4pt}
\small \begin{tabular}{c||S|S|S|S|S|S}
    \multirow{2}{*}{\textbf{Method}} & \multicolumn{2}{c|}{\textbf{Lowercase}} & \multicolumn{2}{c|}{\textbf{Uppercase}} & \multicolumn{2}{c}{\textbf{Combined}} \\
    & {\textbf{WD}} & {\textbf{WI}} & {\textbf{WD}} & {\textbf{WI}} & {\textbf{WD}} & {\textbf{WI}} \\ \hline
    \multirow{1}{*}{Frequentist} & \textbf{84.62} & 76.85 & 89.89 & \textbf{83.01} & 70.50 & 64.13 \\
    \multirow{1}{*}{\cite{ott}} & TCN & TCN & TCN & TCN & TCN & LSTM \\ \hline
    \multirow{2}{*}{SWAG} & 84.44 & \textbf{76.96} & 87.58 & 82.21 & 72.54 & \textbf{66.12} \\
    & TCN & TCN & TCN & TCN & TCN & TCN \\ \hline
    \multirow{1}{*}{Deep} & 83.43 & 73.41 & \textbf{90.31} & 81.26 & \textbf{75.51} & 64.21 \\
    \multirow{1}{*}{Ensembles} & BiLSTM & TCN & TCN & TCN & TCN & TCN \\
\end{tabular}
\caption{Accuracies (in \%) for models trained on \textit{right-handed} writers data and evaluated on \textit{right-handed} writers data. Second row shows the respective model. \textbf{Bold}: best results.}
\label{tab:acc_right}
\end{table}

\newcommand\subfigsized{0.40}
\newcommand\subfigsizedsub{0.49}
\begin{figure*}[t!]
	\centering
	\begin{minipage}[b!]{\subfigsized\linewidth}
    	\begin{minipage}[b!]{\subfigsizedsub\linewidth}
            \centering
            \includegraphics[trim=7 9 7 7, clip, width=1.0\linewidth]{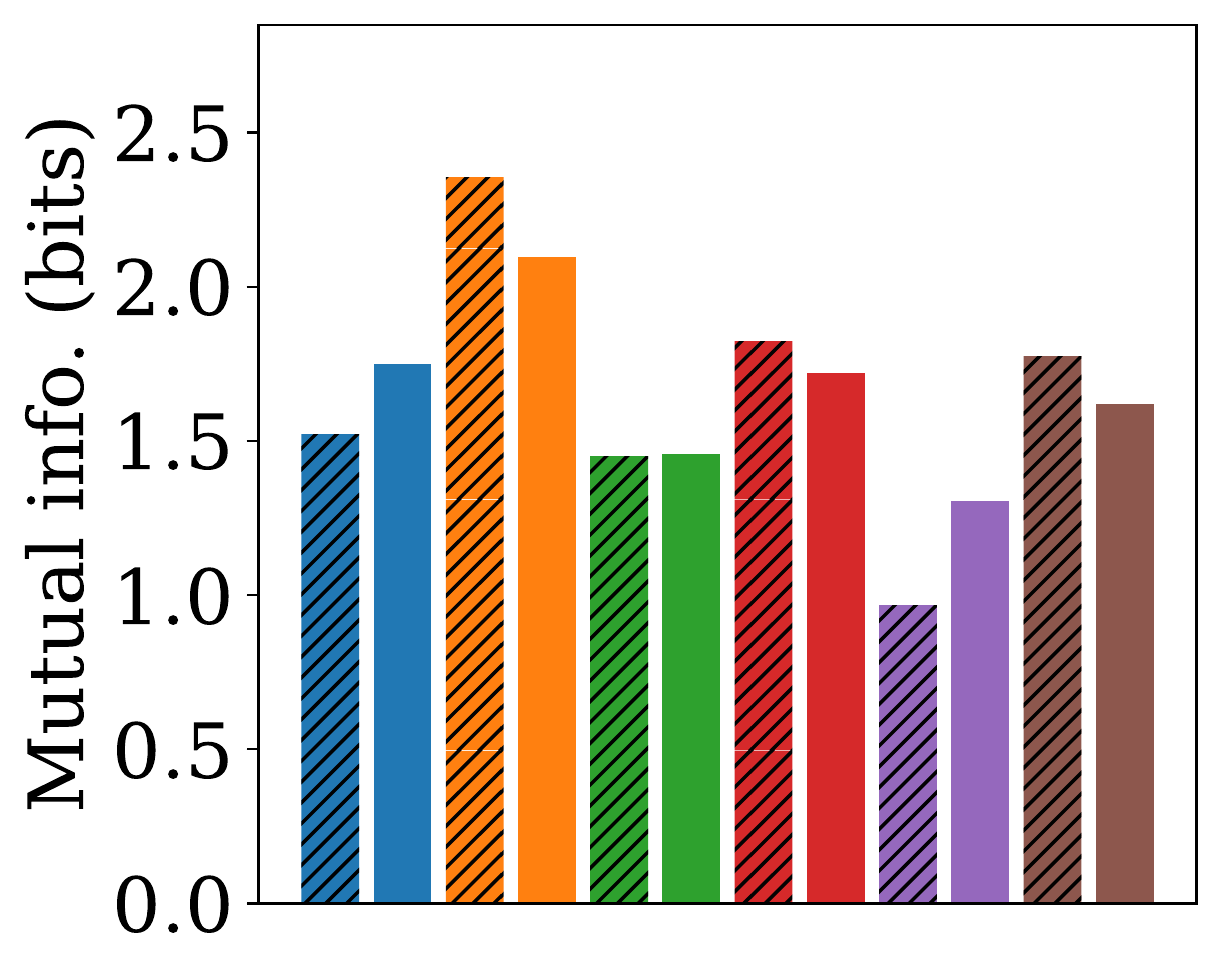}
        \end{minipage}
        \hfill
    	\begin{minipage}[b!]{\subfigsizedsub\linewidth}
            \centering
            \includegraphics[trim=7 9 7 7, clip, width=1.0\linewidth]{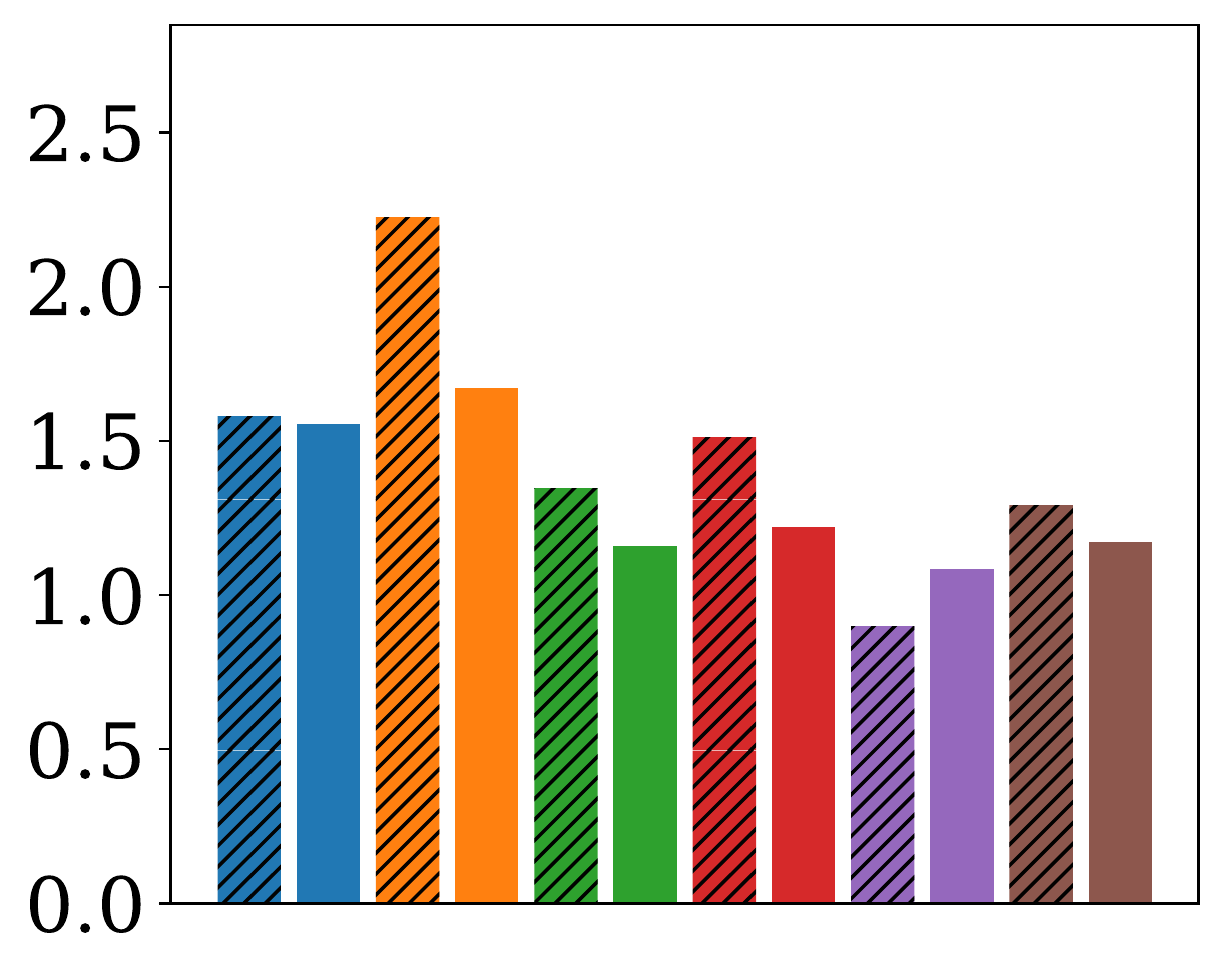}
        \end{minipage}
        \subcaption{Mutual information.}
        \label{fig:letters_mi}
    \end{minipage}
    \hfill
    \begin{minipage}[b!]{\subfigsized\linewidth}
        \begin{minipage}[b!]{\subfigsizedsub\linewidth}
            \centering
            \includegraphics[trim=7 9 7 7, clip, width=1.0\linewidth]{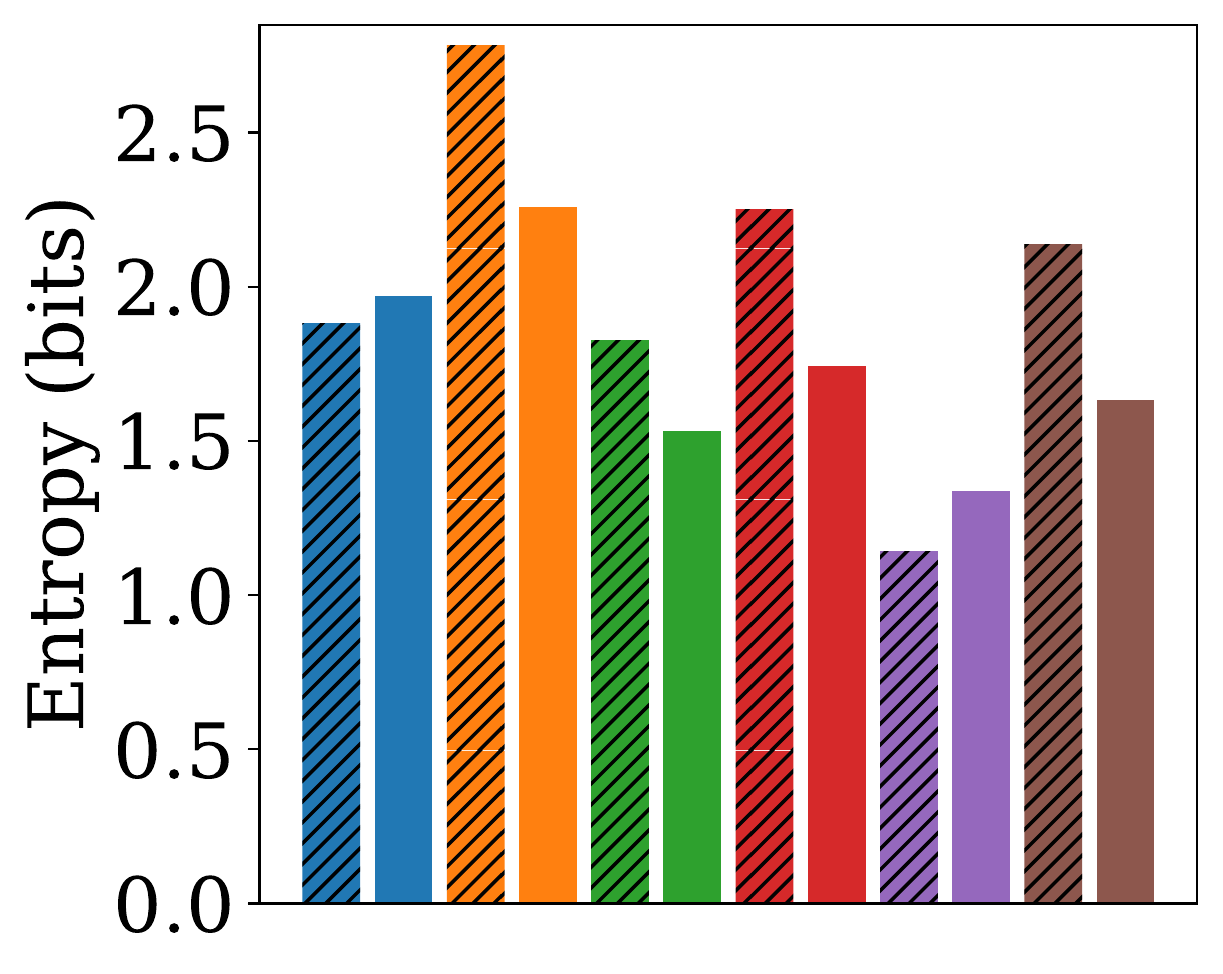}
        \end{minipage}
        \hfill
        \begin{minipage}[b!]{\subfigsizedsub\linewidth}
            \centering
            \includegraphics[trim=7 9 7 7, clip, width=1.0\linewidth]{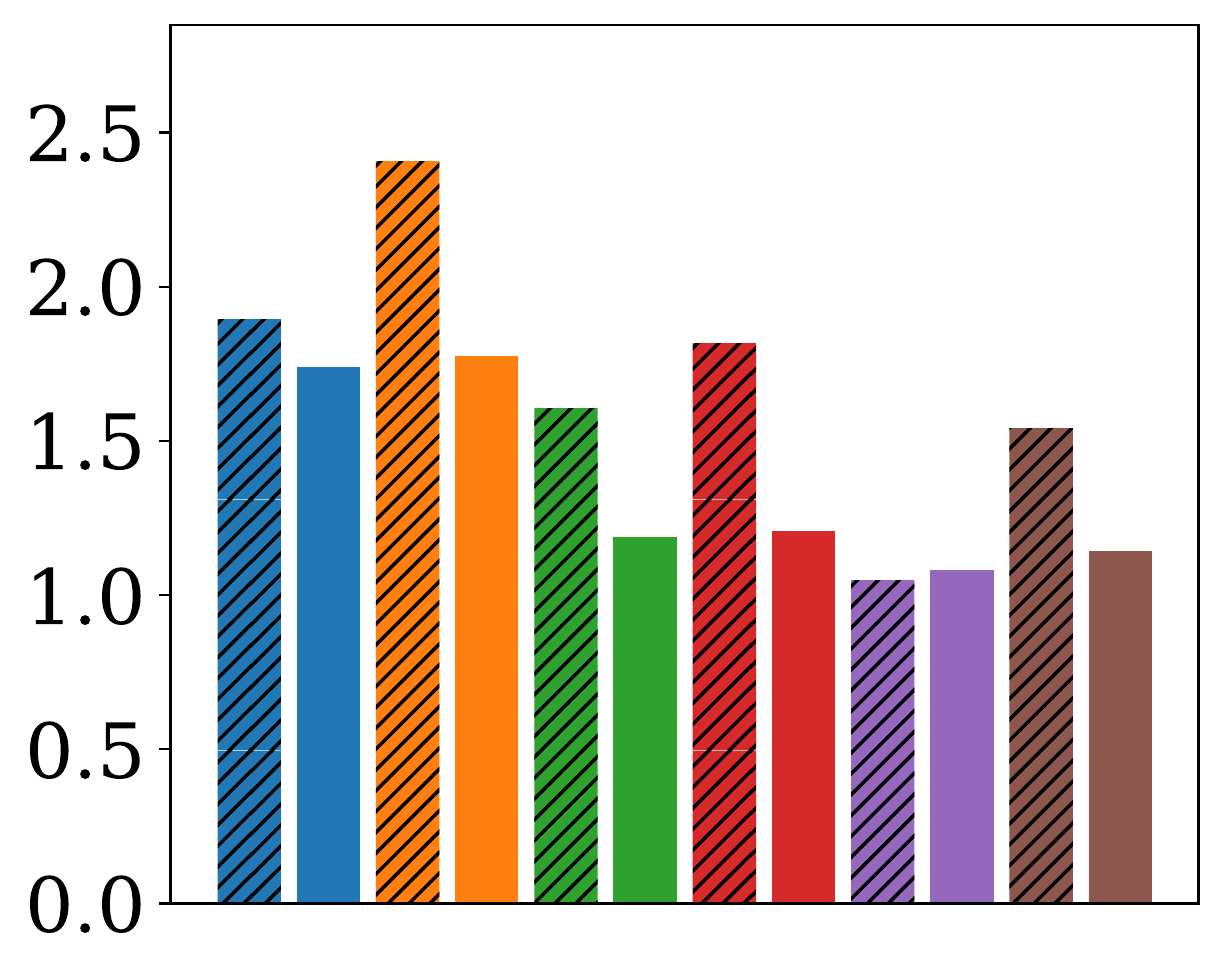}
        \end{minipage}
        \subcaption{Entropy.}
        \label{fig:letters_en}
    \end{minipage}
    \hfill
    \begin{minipage}[t!]{0.19\linewidth}
        \centering
        \includegraphics[trim=55 9 55 9, clip, width=1.0\linewidth]{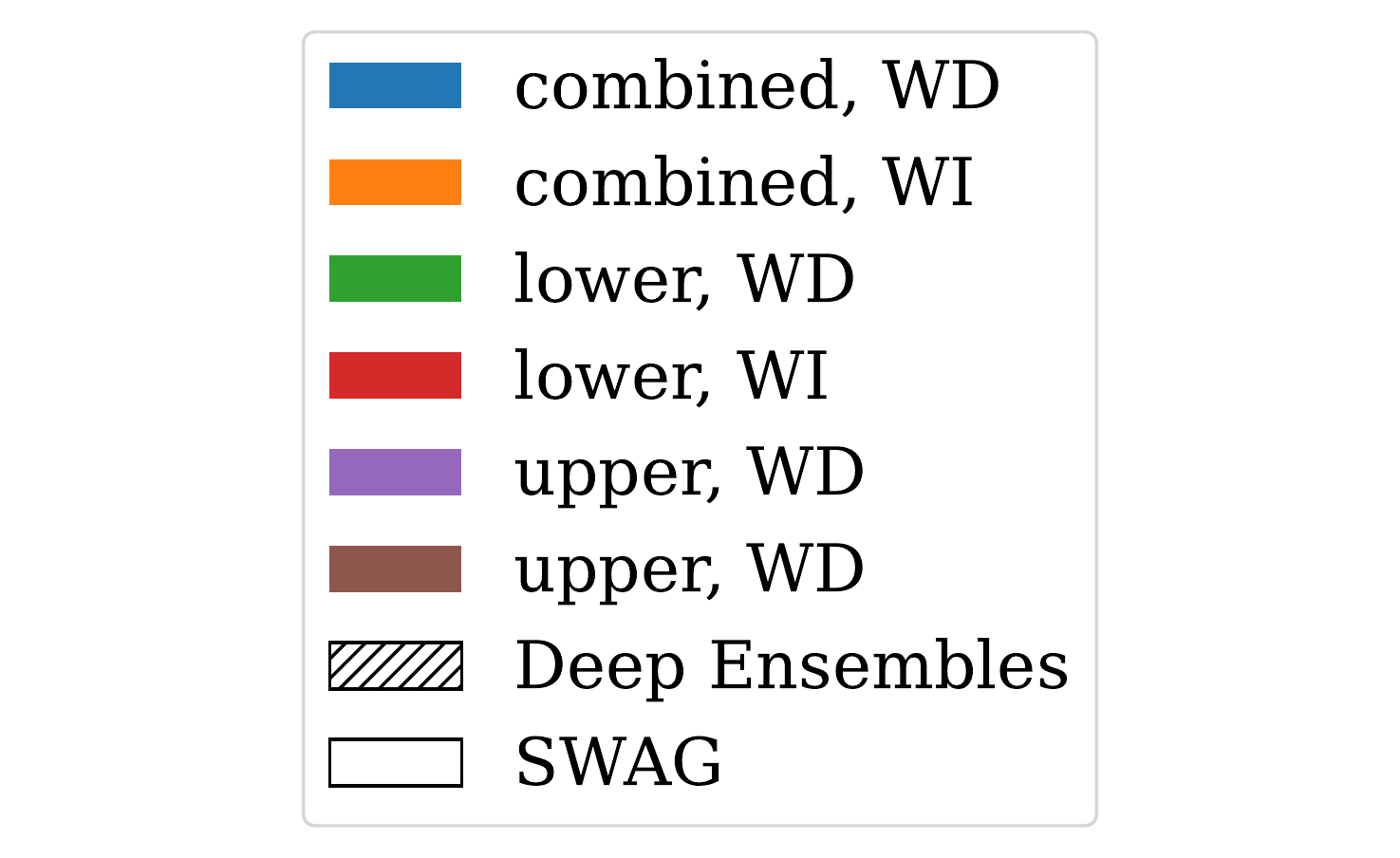}
        \subcaption{Legend.}
        \label{fig:mi_ent_legend}
    \end{minipage}
    \vspace{-0.2cm}
    \caption{Information theoretic uncertainty measures for the Deep Ensemble (dashed) and SWAG (non-dashed) CNN+TCN models. The models are trained on the combined right- and left-handed writers datasets (a and b, left) or only the right-handed writers dataset (a and b, right), and evaluated on the left-handed writers dataset. We provide results for lowercase, uppercase and combined classification tasks.}
    \label{fig:info_left}
    \vspace{-0.1cm}
\end{figure*}

In the following, we summarize the main results. In general, the models perform better on WD classification tasks than on WI tasks. Architectures with TCN units outperform LSTM and BiLSTM units on most tasks.

\paragraph{Evaluation on Handedness (trained on right-handed writers).} SWAG and Deep Ensemble models perform very similarly to frequentist models proposed in \cite{ott} in terms of predicitve accuracy (see Table~\ref{tab:acc_right}), being at most 3\% points below and 5\% points above a respective frequentist model. When applying models trained with right-handed data on the left-handed datasets, the performance ranges from 33.27\% to 49.87\% accuracy (see Table~\ref{tab:acc_split}) which is substantially better than ``pure guessing'' -- our models make informed decisions after shifting domains, albeit at a lower standard. A possible reason is that certain sensors produce nearly identical signals regardless of the orientation of the pen. For example, the accelerometer at the bottom of the pen should give the same readings for left-handed writers when writing \texttt{"I"} and \texttt{"i"} as for right-handed writers, since it is simply a downward motion regardless of the writing hand.

\begin{table}[t!]
\centering
\setlength{\tabcolsep}{1.9pt}
\small \begin{tabular}{c|c||S|S|S|S|S|S}
    \multirow{2}{*}{\textbf{Method}} & & \multicolumn{2}{c|}{\textbf{Lowercase}} & \multicolumn{2}{c|}{\textbf{Uppercase}} & \multicolumn{2}{c}{\textbf{Combined}} \\
    & & {\textbf{WD}} & {\textbf{WI}} & {\textbf{WD}} & {\textbf{WI}} & {\textbf{WD}} & {\textbf{WI}} \\ \hline
    \multirow{2}{*}{SWAG}& right & 83.73 & 76.27 & 87.10 & 81.69 & 72.13 & 65.41 \\
    & left & \textbf{55.51} & \textbf{45.91} & 55.04 & \textbf{50.67} & \textbf{46.08} & \textbf{39.26} \\ \hline
    \multirow{2}{*}{Deep Ensembles} & right & 83.07 & 73.87 & 89.92 & 80.86 &  75.29 & 64.22 \\
    & left & 45.25 & 37.00 & \textbf{62.73} & 48.31 & 45.95 & 33.27 \\ \hline
    \multirow{1}{*}{Best BNN Method} & right & \textbf{84.44} & \textbf{76.96} & \textbf{90.31} & \textbf{82.21} & \textbf{75.51} & \textbf{66.12} \\
    \multirow{1}{*}{(right-handed)} & left & 42.55 & 44.19 & 49.87 & 48.54 & 33.68 & 36.20 \\
\end{tabular}
\caption{Accuracies (in \%) for best models trained on right- and left-handed data and evaluated on \textit{right-handed or left-handed} writers data \textit{separately}, compared to the best performing models which were only trained on right-handed data. \textbf{Bold}: best results.}
\label{tab:acc_split}
\end{table}

\begin{table}[t!]
\centering
\setlength{\tabcolsep}{2.4pt}
\small \begin{tabular}{c||S|S|S|S|S|S}
    \multirow{2}{*}{\textbf{Method}} & \multicolumn{2}{c|}{\textbf{Lowercase}} & \multicolumn{2}{c|}{\textbf{Uppercase}} & \multicolumn{2}{c}{\textbf{Combined}} \\
    & {\textbf{WD}} & {\textbf{WI}} & {\textbf{WD}} & {\textbf{WI}} & {\textbf{WD}} & {\textbf{WI}} \\ \hline
    \multirow{2}{*}{SWAG} & \textbf{81.85} & \textbf{74.24} & 84.92 & \textbf{79.58} & 70.37 & \textbf{63.64} \\
    & TCN & TCN & TCN & TCN & TCN & TCN \\ \hline
    \multirow{1}{*}{Deep} & 80.55 & 71.41 & \textbf{88.07} & 78.65 & \textbf{73.31} & 62.14 \\
    \multirow{1}{*}{Ensembles} & LSTM & TCN & TCN & TCN & TCN & TCN \\
\end{tabular}
\caption{Accuracies (in \%) for models trained on \textit{right- and left-handed} writers data and evaluated on \textit{right-handed} writers data. Second row shows the respective model. \textbf{Bold}: best results.}
\label{tab:left_acc}
\end{table}

\newcommand\subfigsizea{0.244}
\begin{figure*}[t!]
	\centering
	\begin{minipage}[b!]{\subfigsizea\linewidth}
        \centering
        \includegraphics[trim=5 7 7 7, clip, width=1.0\linewidth]{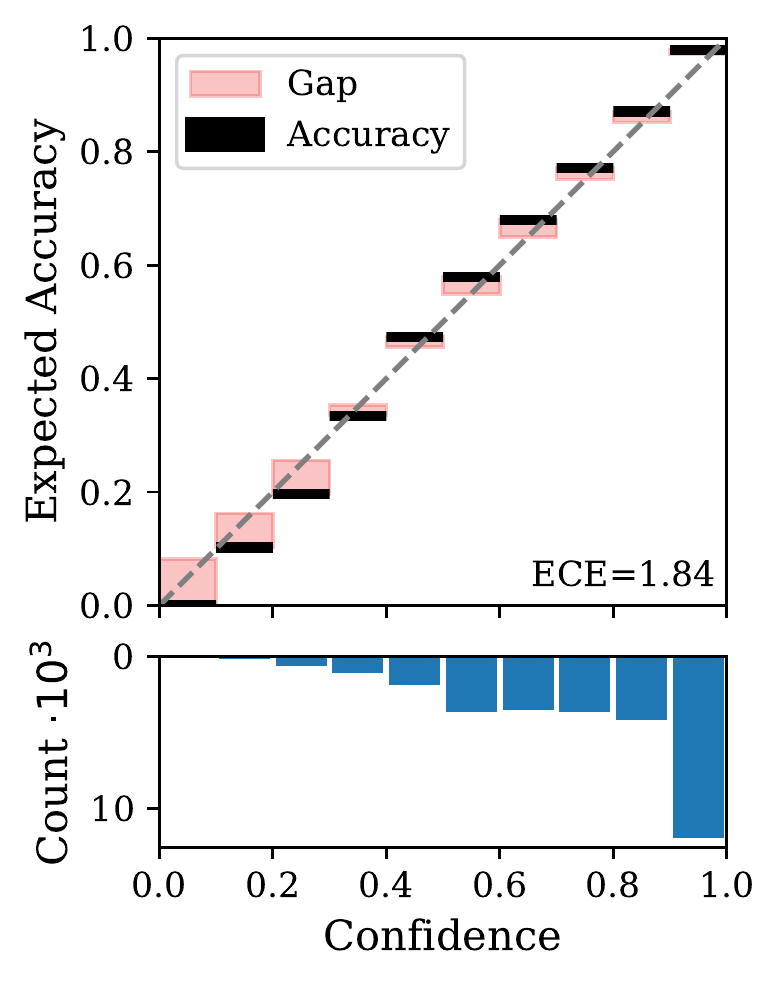}
        \subcaption{Evaluated on right-handed writers data.}
        \label{fig:eval_rel_dia_a}
    \end{minipage}
    \hfill
	\begin{minipage}[b!]{\subfigsizea\linewidth}
        \centering
        \includegraphics[trim=5 7 7 7, clip, width=1.0\linewidth]{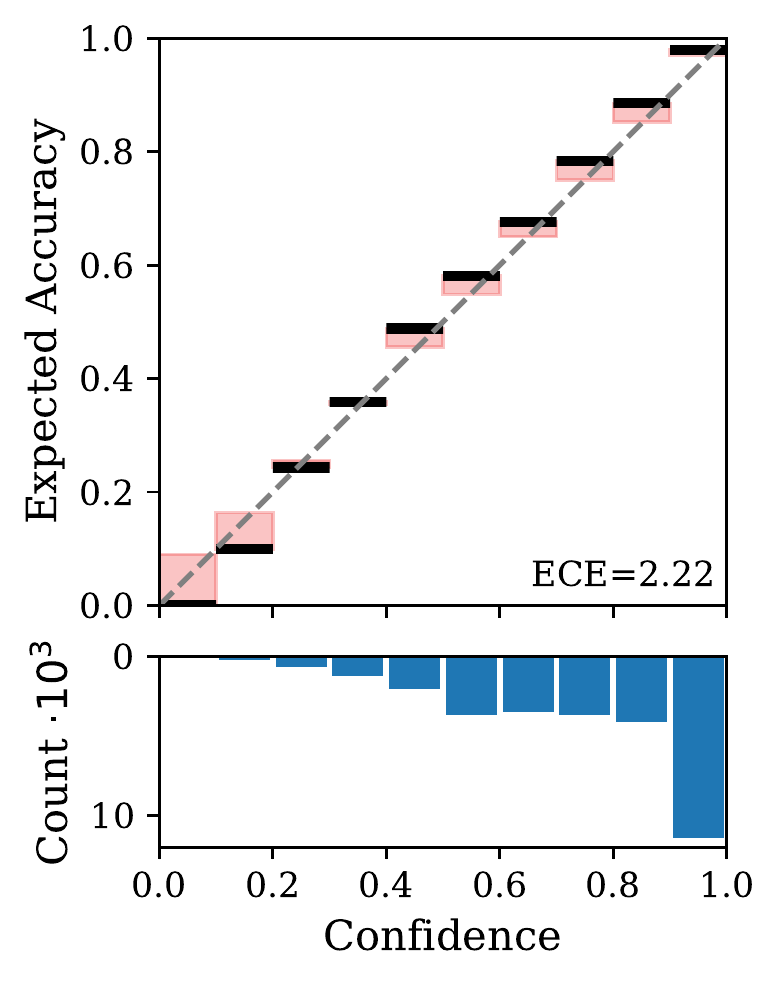}
        \subcaption{Evaluated on right-handed writers data.}
        \label{fig:eval_rel_dia_c}
    \end{minipage}
    \hfill
	\begin{minipage}[b!]{\subfigsizea\linewidth}
        \centering
        \includegraphics[trim=5 7 7 7, clip, width=1.0\linewidth]{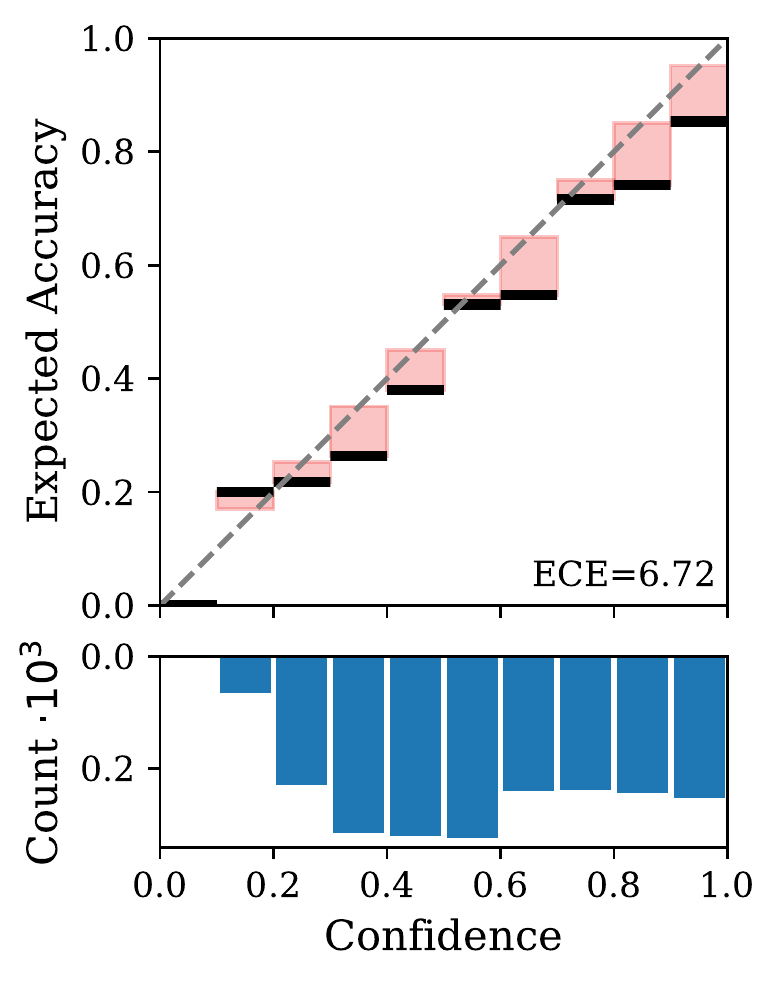}
        \subcaption{Evaluated on left-handed writers data.}
        \label{fig:eval_rel_dia_b}
    \end{minipage}
    \hfill
	\begin{minipage}[b!]{\subfigsizea\linewidth}
        \centering
        \includegraphics[trim=5 7 7 7, clip, width=1.0\linewidth]{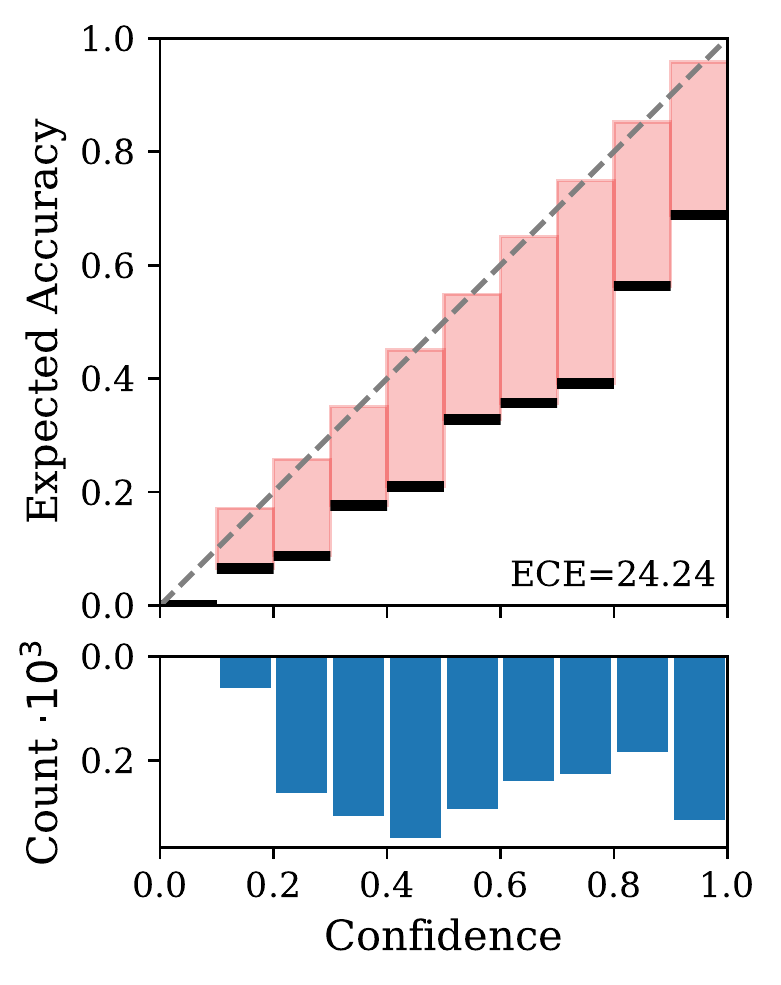}
        \subcaption{Evaluated on left-handed writers data.}
        \label{fig:eval_rel_dia_d}
    \end{minipage}
    \vspace{-0.2cm}
    \caption{Reliability diagram for the Deep Ensemble CNN+TCN model trained on the combined WD datasets. a) and c): Trained on the combined right- and left-handed writers datasets. b) and d): Trained on right-handed writers only.}
\label{fig:cal1}
\end{figure*}

\paragraph{Evaluation on Handedness (trained on right- and left-handed writers).} When evaluating performance on right-handed data, models trained only on right-handed datasets consistently outperform models trained on both datasets combined and yield between 2\% points and 12\% points higher accuracies (see Table~\ref{tab:left_acc}). This performance loss is compensated by a performance gain for left-handed data. Still, the performance is not up to par with right-handed data; this gap may be due to a ``writing style'' particular to every writer that especially influences the gyroscope and magnetometer measurements. More importantly, left-handed writers have a writing style different to right-handed writers which is perhaps exactly what the right-handed models never learned in order to address the style of left-handed writers, underlining the need for a sufficient amount of samples to get a good representation of various writing styles.

\paragraph{Analysis of Uncertainty.} Figure~\ref{fig:info_left} shows the MI and entropy for SWAG and Deep Ensemble models evaluated on the left-handed data. The barplots show that the models trained on only right-handed data display lower uncertainty (i.e., higher confidence) compared to models trained on combined data. However, this higher confidence is not empirically justified when looking at the reliability diagrams in Figure~\ref{fig:cal1}, which point out that models trained without left-handed writers data are miscalibrated and therefore overconfident. Models trained on the combined writers (Figures~\ref{fig:eval_rel_dia_a} and \ref{fig:eval_rel_dia_b}) provide more realistic accuracies when applied to the left-handed data (ECE of 6.72). The ECE is even higher (24.24) for left-handed evaluation without left-handers in the training set (see Figure~\ref{fig:eval_rel_dia_d}). For a separate evaluation for each character, see Appendix~\ref{sec_app_character}.

\subsection{Uncertainty Analysis based on [Kwon et al.]}\label{analysis_kwon}

In Figure~\ref{fig:both} we visualize the aleatoric and epistemic uncertainty as well as the confusion matrix for the Deep Ensemble model and the combined task. For SWAG model results, see Appendix~\ref{app_swag_results}. In the aleatoric uncertainty heatmap (Figure~\ref{fig:alea_both}) we observe a trace with negative values at the lower end of the scale. Note that for off-diagonal values, the aleatoric uncertainty is higher for lower softmax values. Here, two softmax outputs (with the highest values) coincide on average (see Section~\ref{sub_sectin_ud_knwon}). This means that the model tends to confuse the two classes. The most prominent off-diagonal strip corresponds to the upper- and lowercase pairs. This makes intuitive sense since, e.g., the lowercase \texttt{"u"} and uppercase \texttt{"U"} are written similarly. This effect is consequently not present for less similar pairs like \texttt{"a"} and \texttt{"A"}. We can see this effect also for \texttt{"l"} (lowercase \texttt{"L"}) and \texttt{"I"} (uppercase \texttt{"i"}). A very similar pattern can also be observed in the confusion matrix (see Figure~\ref{fig:conf_both}), confirming that the trained model is not only unsure about how to classify these pairs, but is also empirically worse in the respective classification task.

\newcommand\subfigsizeb{0.33}
\begin{figure*}[t!]
\centering
\begin{subfigure}{\subfigsizeb\textwidth}
    \centering
    \includegraphics[width=1.0\linewidth]{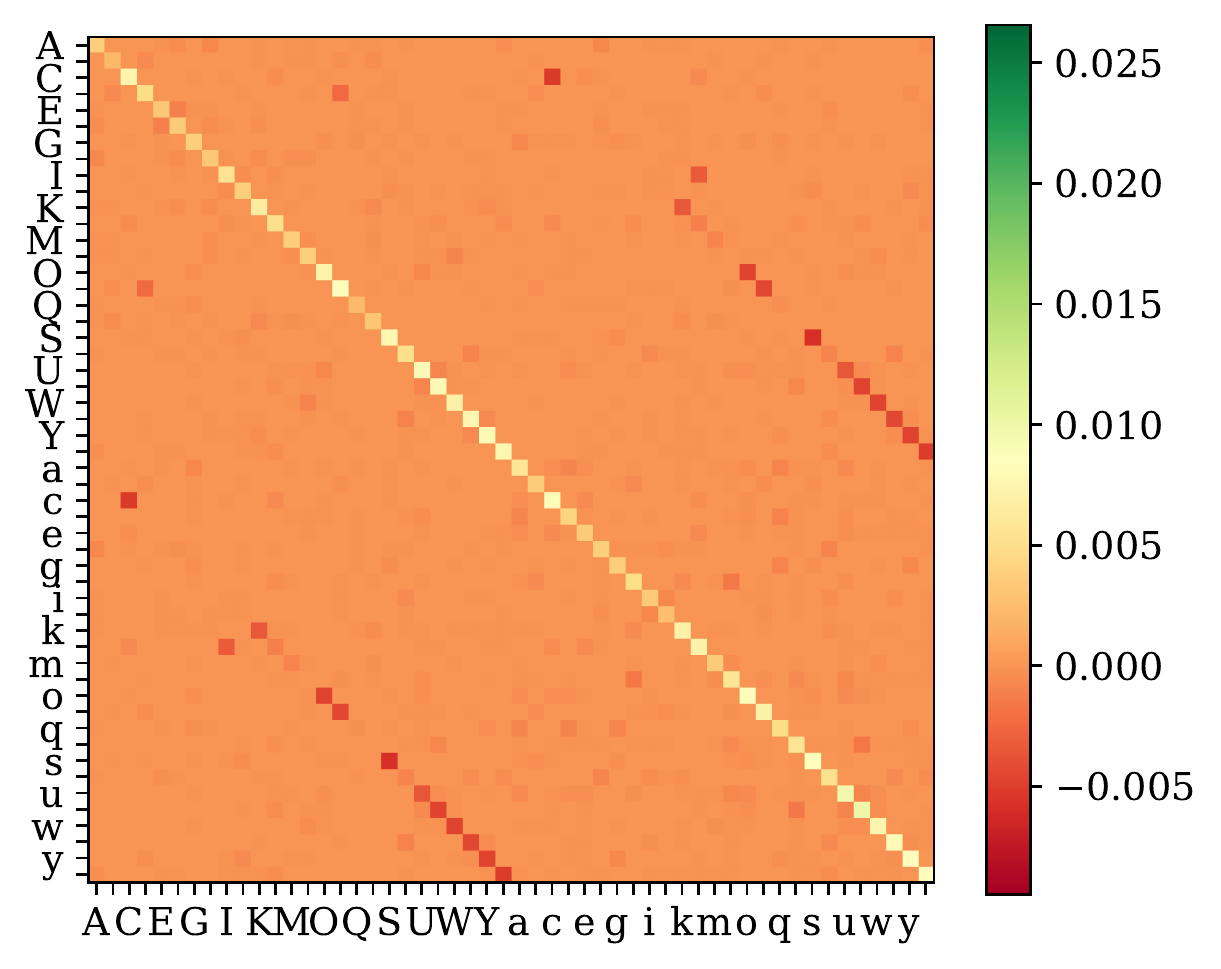}
    \vspace{-0.5cm}
    \caption{Aleatoric uncertainty.}
    \label{fig:alea_both}
\end{subfigure}
\hfill
\begin{subfigure}{\subfigsizeb\textwidth}
    \centering
    \includegraphics[width=1.0\linewidth]{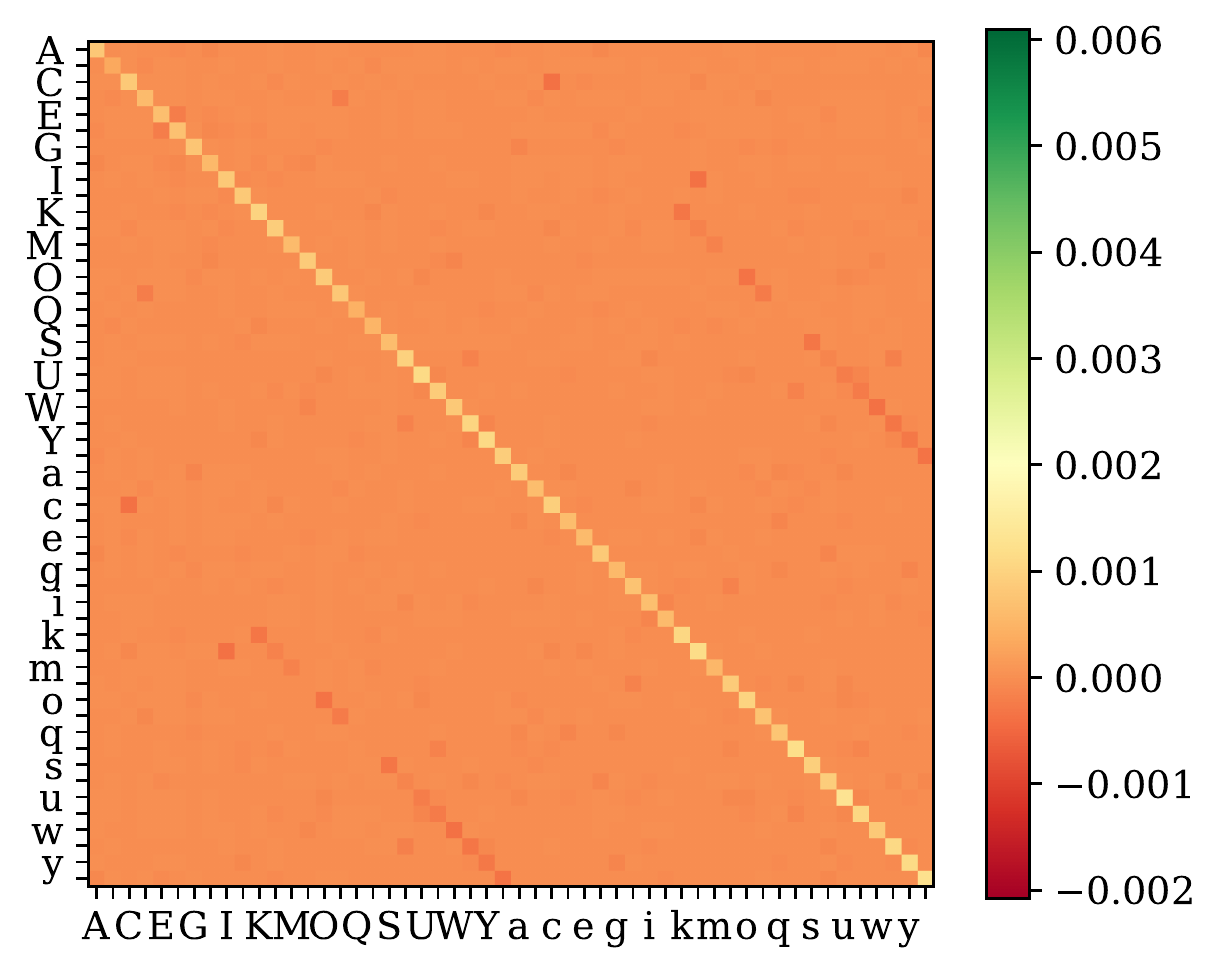}
    \vspace{-0.5cm}
    \caption{Epistemic uncertainty.}
    \label{fig:epis_both}
\end{subfigure}
\hfill
\begin{subfigure}{\subfigsizeb\textwidth}
    \centering
    \includegraphics[width=1.0\linewidth]{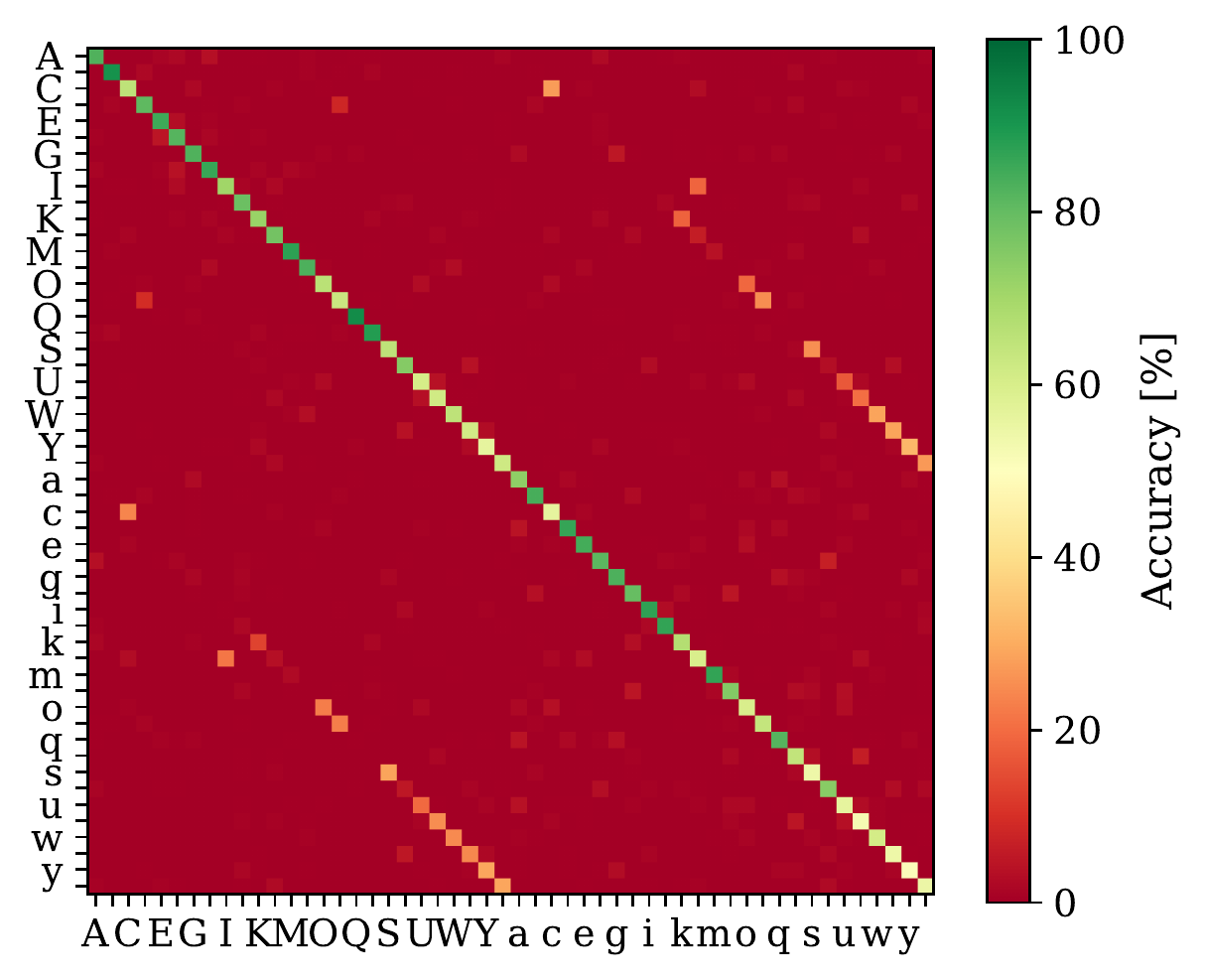}
    \vspace{-0.5cm} 
    \caption{Confusion matrix of accuracy.}
    \label{fig:conf_both}
\end{subfigure}
\vspace{-0.3cm}
\caption{Uncertainty prediction for the Deep Ensemble CNN+TCN model trained on the combined WD (right-handed only) dataset. Note that the color scale is fixed for all subplots for comparability with Figure~\ref{fig:uppercase_conf}, and that we skipped every second character label for readability.}
\label{fig:both}
\vspace{-0.1cm}
\end{figure*}

\begin{figure*}[t!]
\centering
\begin{subfigure}{\subfigsizeb\textwidth}
    \centering
    \includegraphics[width=1\linewidth]{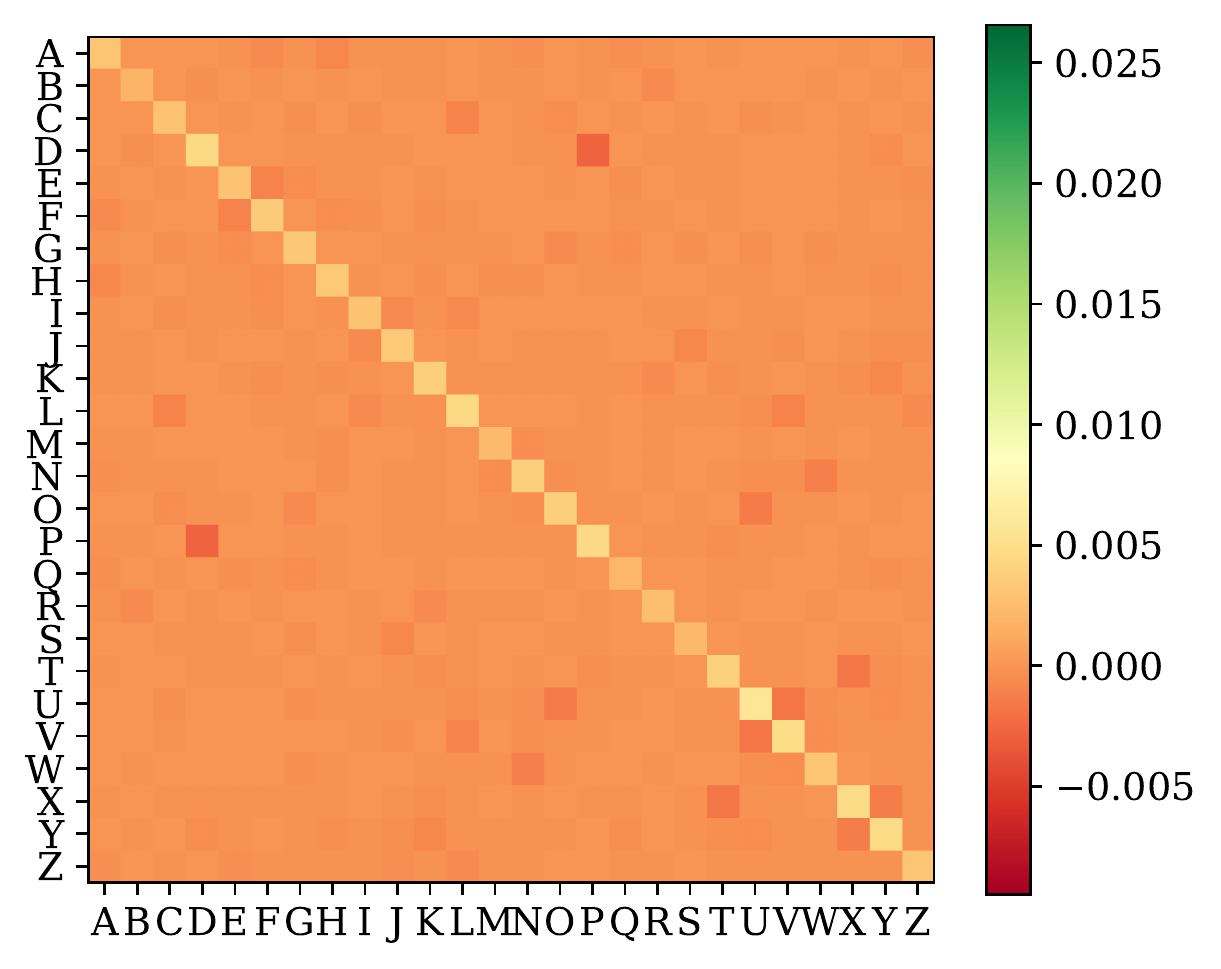}
    \vspace{-0.5cm}
    \caption{Aleatoric uncertainty.}
    \label{fig_app_alea_upp}
\end{subfigure}
\hfill
\begin{subfigure}{\subfigsizeb\textwidth}
    \centering
    \includegraphics[width=1\linewidth]{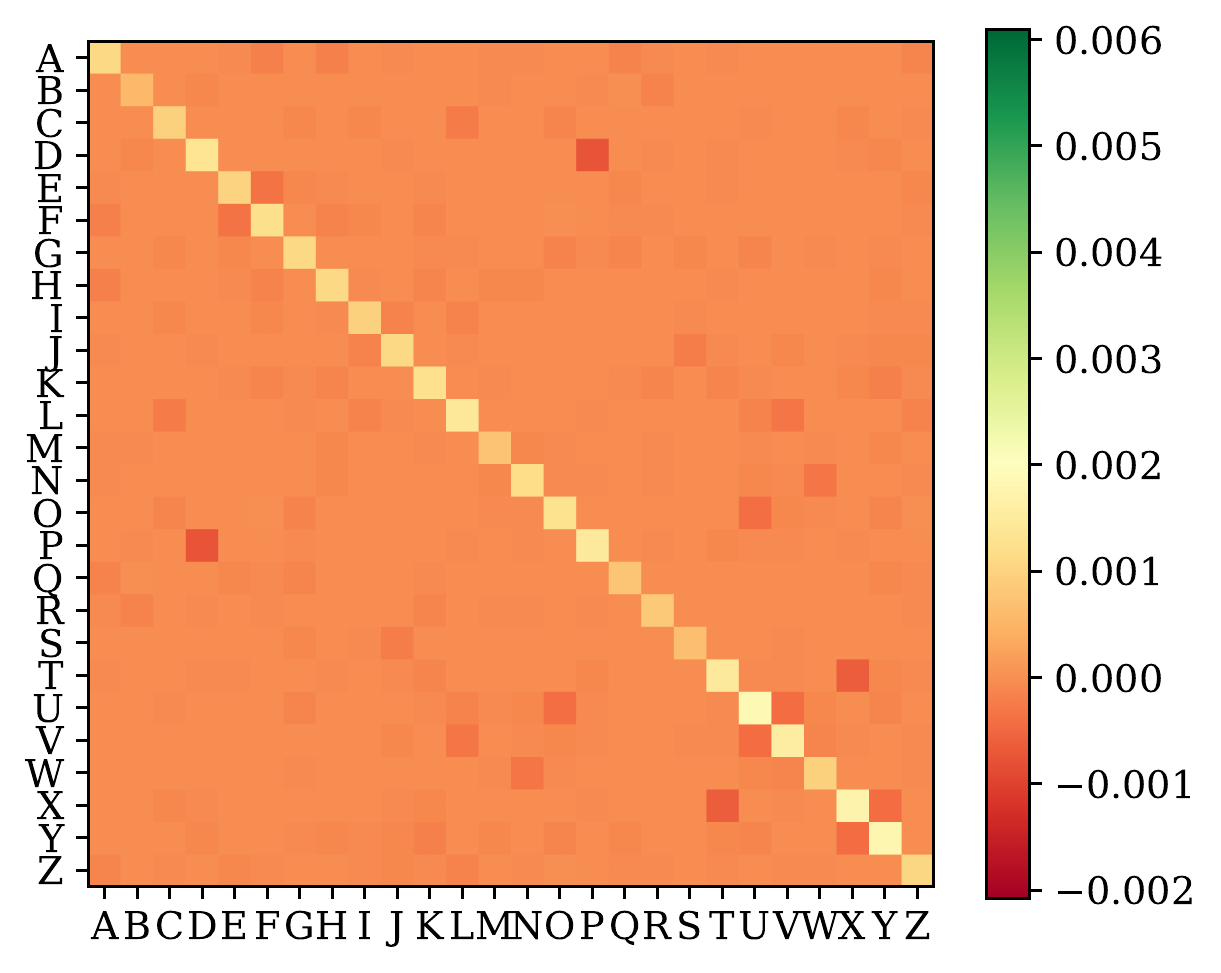} 
    \vspace{-0.5cm}
    \caption{Epistemic uncertainty.}
    \label{fig_app_epis_upp}
\end{subfigure}
\hfill
\begin{subfigure}{\subfigsizeb\textwidth}
    \centering
    \includegraphics[width=1\linewidth]{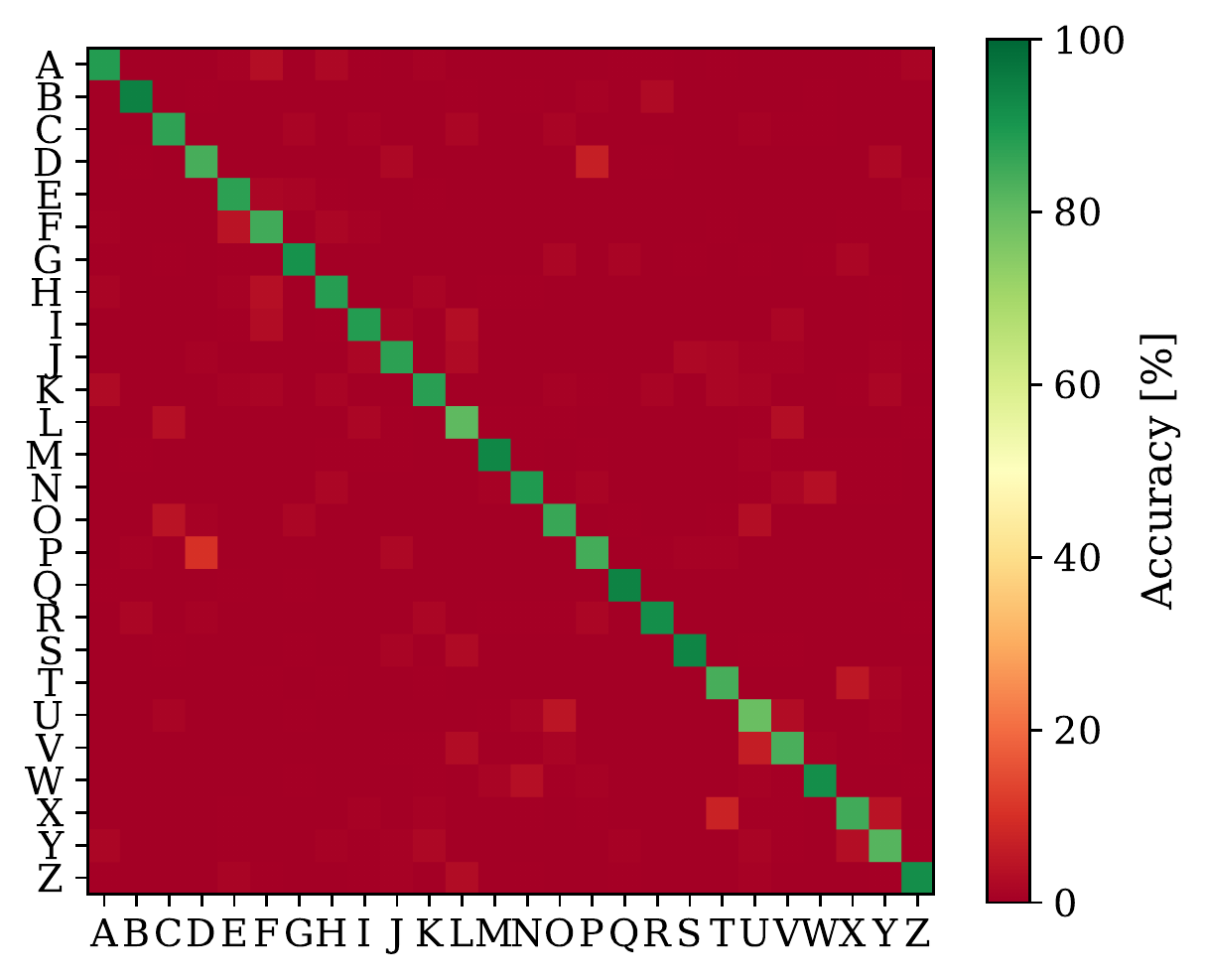} 
    \vspace{-0.5cm}
    \caption{Confusion matrix of accuracy.}
    \label{fig_app_conf_upp}
\end{subfigure}
\vspace{-0.3cm}
\caption{Uncertainty prediction for the Deep Ensemble CNN+TCN model trained on the uppercase WD (right-handed only) dataset. Note that the color scale is fixed for all subplots for comparability with Figure~\ref{fig:both} and \ref{fig:lowercase_conf}.}
\label{fig:uppercase_conf}
\vspace{-0.1cm}
\end{figure*}

These patterns allow for further interesting insights. For example, one might expect this pattern to occur for \texttt{"i"} and \texttt{"j"}, but the corresponding heatmap entries lack signs of confusion of the model. Similarity between characters consequently hinges on the similarity of \textit{motions} while writing. Two characters with small differences are written similarly but in different \textit{sizes}. This also holds for specific parts of the letters. For example, \texttt{"n"} and \texttt{"h"} have a higher aleatoric uncertainty in Figure~\ref{fig:alea_both}; the major difference being that one tiny part of \texttt{"h"} is longer.

Somewhat puzzling is that we see the same effect in the epistemic uncertainty heatmap (see Figure~\ref{fig:epis_both}), where such pairs with high similarity lead to negative values. 
When one entry of the softmax output values is below and another entry above the respective sample mean, negative epistemic uncertainty is implied. This leads to some kind of discriminative power due to the negative ``covariance'' for which there is little justification. We thus advise caution when interpreting the epistemic uncertainty in this context.

\subsection{Uncertainty based on Information Theory}\label{section_eval_uc_kwon}

\begin{figure}[t!]
	\centering
	\begin{minipage}[b]{0.51\linewidth}
        \centering
        \includegraphics[trim=0 10 0 6, clip, width=1.0\columnwidth]{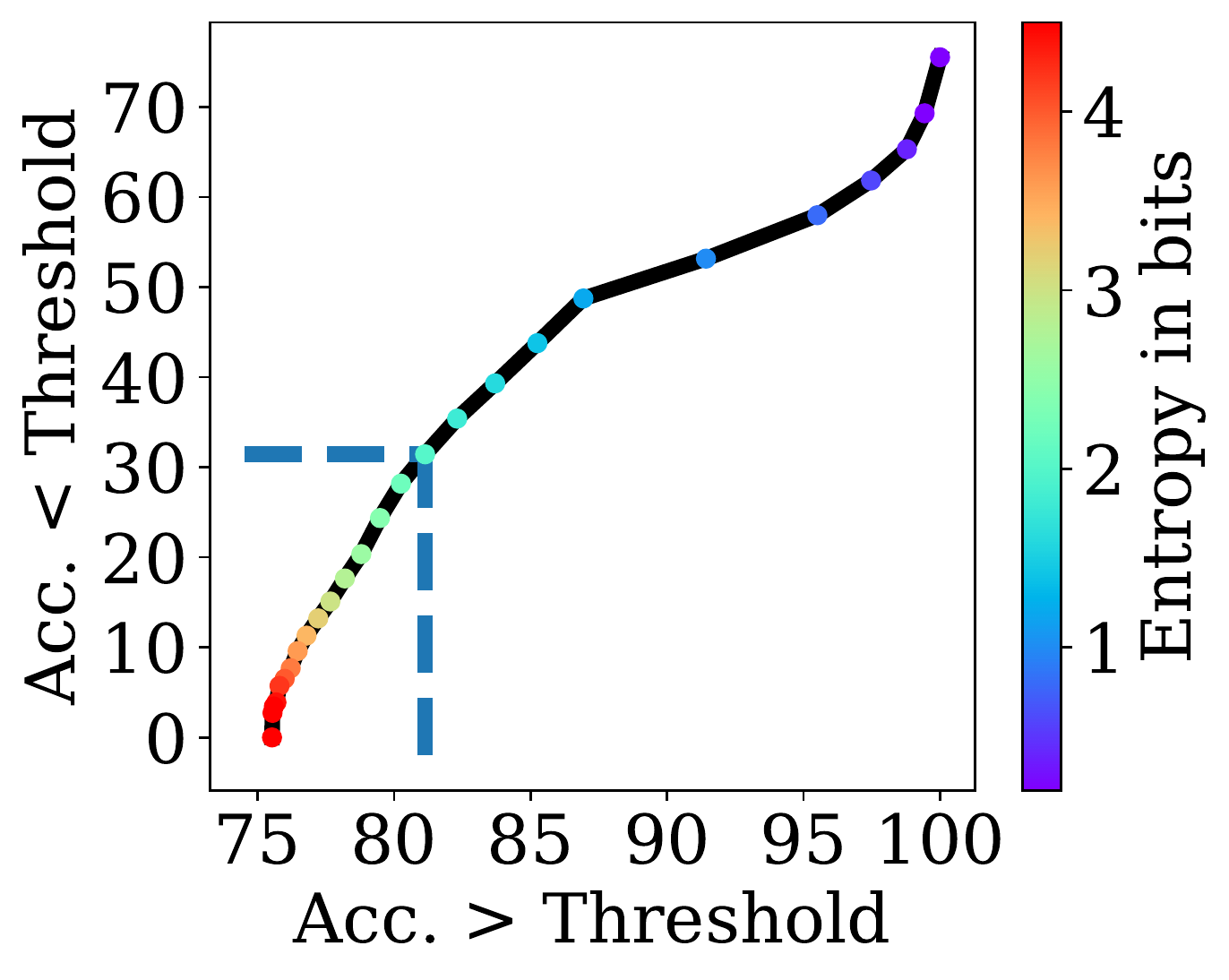}
        \subcaption{Sample accuracies below and above an entropy threshold.}
        \label{fig:trade-off}
    \end{minipage}
    \hfill
	\begin{minipage}[b]{0.47\linewidth}
        \centering
        \includegraphics[trim=0 8 0 6, clip, width=1.0\columnwidth]{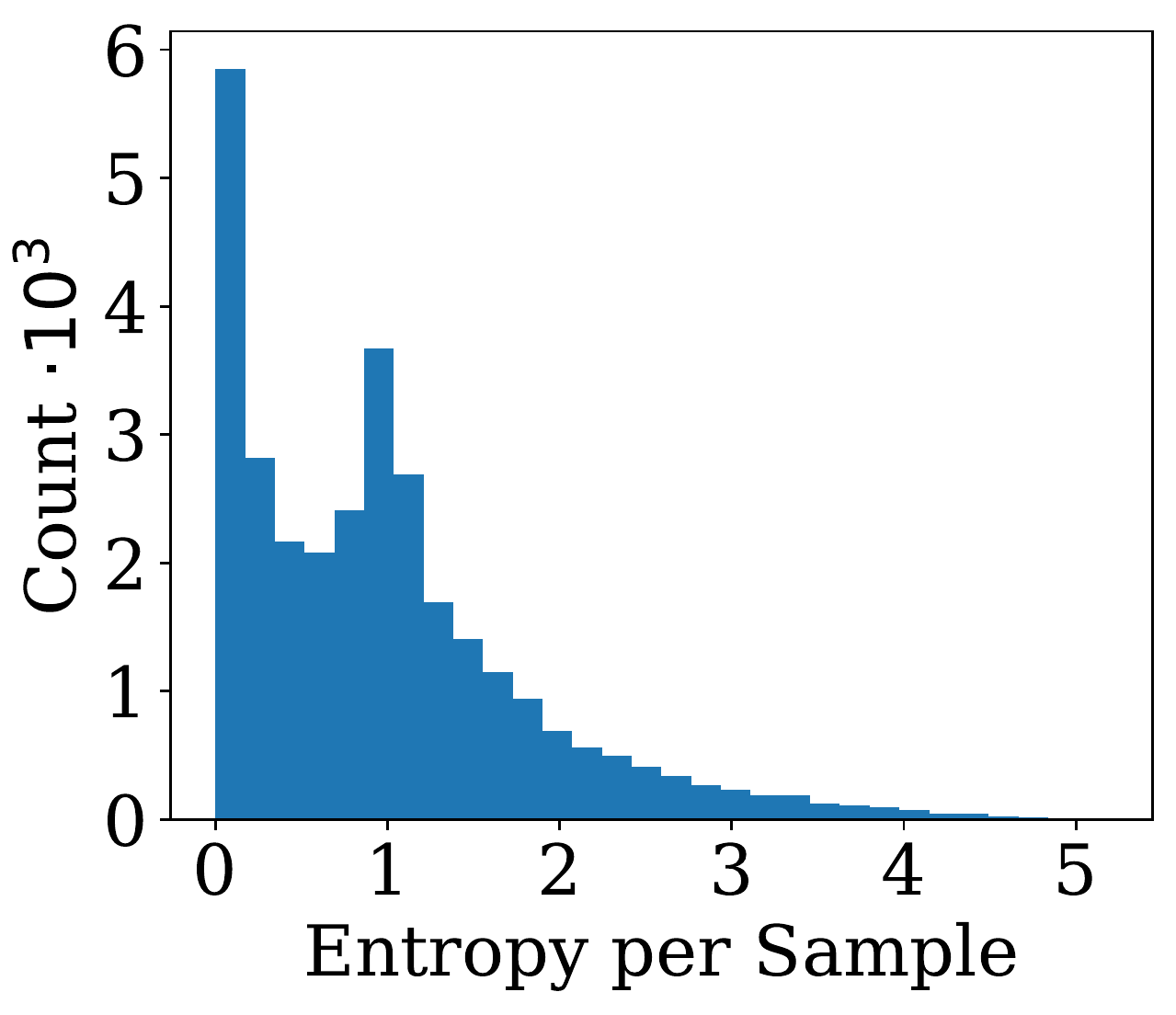}
        \subcaption{Histogram visualizing the entropy distribution.}
        \label{fig:hist_en}
    \end{minipage}
    \caption{Accuracy and entropy for the Deep Ensemble CNN+TCN model trained on the combined WD (right-handed only) dataset.}
    \label{fig:cutoff}
\end{figure}

We further highlight the trade-off when using information theory-based measures to decide whether a sample is too uncertain to classify correctly. This is depicted by Figure~\ref{fig:trade-off} showing the relationship between classification accuracies and different threshold values. We choose the entropy as the target metric for uncertainty evaluation (MI would work analogously). On the x-axis is the accuracy of the samples above the threshold, i.e., samples our model feels confident about classifying correctly. On the y-axis is the accuracy for the samples below the threshold. These values would be considered as too inaccurate to confidently classify. Setting the threshold to 2.0 bits would approximately yield an accuracy of 82\% for the observations above this threshold and approx.~31\% accuracy for observations below this threshold (emphasized by the dashed lines). Figure~\ref{fig:hist_en} depicts the entropy distribution and further clarifies this point. Convincingly, the accuracy reduces to almost zero for very high entropy samples. Note that the accuracy does not need to decrease with an increasing entropy threshold or even be zero for very high entropy values, even though this is generally true for our models.

\subsection{Summary and Limitations}
\label{section_eval_limitations}

\paragraph{Uncertainty Decomposition.} Neither uncertainty quantification method shows notable differences between aleatoric and epistemic uncertainty. The heatmaps exhibit the same ``strip'' for similar characters and give no hints to different sources of uncertainty (data-driven or systemic confusion). The benefits of this kind of uncertainty differentiation are limited, but measuring the total uncertainty can still be useful for domain adaptation or the detection of wrong labels.

\paragraph{Real-World Link.} Since the models trained on right- and left-handed writers lead to lower data confidence compared to models trained only on right-handed writers (see Figure~\ref{fig:info_left}), it is unclear how well the measured MI and entropy translate to the real-world uncertainty. Therefore, verifying uncertainty remains a limitation in our interpretation. While we can discriminate between the entropy associated with different samples, pre-defining thresholds for uncertain samples is challenging due to the following reasons: (1) Raw sensor data is elaborate to interpret and making statements about, e.g., the writing style from sensor data is hardly possible -- which, in turn, is connected to model uncertainty. (2) Interpreting the \textit{graphomotoricity} qualitatively, e.g., for teaching hand writing, a qualified expert in this field is required. (3) Different writing domains (different pens, surfaces etc.) lead to different requirements for the uncertainty threshold.

\section{Conclusion}
\label{chap_conclusion}

We employed SWAG and Deep Ensembles for OnHW recognition with left- and right-handed writers, a spatio-temporal MTS classification task with domain shift. We critically evaluated aleatoric and epistemic uncertainty using confidence calibration, ECE and reliability diagrams. In summary, (1) the model performance only partly relates to the handedness of writers, (2) our models are over-confident and miscalibrated when only trained with right-handed writers and evaluated on left-handed writers, (3) the uncertainty of the models for small and capital characters combined is related to lower classification accuracy, and (4) the entropy and mutual information for individual samples correlate well with the accuracy of our models. Our comparison of different ways to decompose uncertainty easily generalizes to other classification tasks and can be useful for spatio-temporal reasoning. In terms of Bayesian inference, SWAG and Deep Ensemble models perform similarly, while SWAG is computationally less expensive.


\section*{Acknowledgements}
This work was supported by the Federal Ministry of Education and Research (BMBF) of Germany by Grant No. 01IS18036A (David R\"ugamer) and by the research program Human-Computer-Interaction through the project "Schreibtrainer", Grant No. 16SV8228, as well as by the Bavarian Ministry for Economic Affairs, Infrastructure, Transport and Technology through the Center for Analytics-Data-Applications (ADA-Center) within the framework of "BAYERN DIGITAL II".


\small
\bibliographystyle{ijcai22}
\bibliography{ijcai22}

\begin{thebibliography}{}

\bibitem[\protect\citeauthoryear{Blundell \bgroup \em et al.\egroup
  }{2015}]{blundell}
Charles Blundell, Julien Cornebise, Koray Kavukcuoglu, and Daan Wierstra.
\newblock {Weight Uncertainty in Neural Network}.
\newblock In {\em \href{https://dl.acm.org/doi/10.5555/3045118.3045290}{ICML}},
  volume~37, pages 1613--1622, July 2015.

\bibitem[\protect\citeauthoryear{Cai \bgroup \em et al.\egroup
  }{2014}]{cai_chen}
Ruichu Cai, Jiawei Chen, Zijian Li, Wei Chen, Keli Zhang, Junjian Ye, Zhuozhang
  Li, Xiaoyan Yang, and Zhenjie Zhang.
\newblock {Time Series Domain Adaptation via Sparse Associative Structure
  Alignment}.
\newblock In {\em
  \href{https://ojs.aaai.org/index.php/AAAI/article/view/16846}{AAAI}}, volume
  216, pages 76--102, 2014.

\bibitem[\protect\citeauthoryear{Daxberger \bgroup \em et al.\egroup
  }{2021}]{daxberger_kristiadi}
Erik Daxberger, Agustinus Kristiadi, Alexander Immer, Runa Eschenhagen,
  Matthias Bauer, and Philipp Hennig.
\newblock {Laplace Redux - Effortless Bayesian Deep Learning}.
\newblock In {\em
  \href{https://proceedings.neurips.cc/paper/2021/file/a7c9585703d275249f30a088cebba0ad-Paper.pdf}{NIPS}},
  December 2021.

\bibitem[\protect\citeauthoryear{Degroot and Fienberg}{1983}]{Degroot1983TheCA}
Morris~H. Degroot and Stephen~E. Fienberg.
\newblock {The Comparison and Evaluation of Forecasters}.
\newblock In {\em \href{https://www.jstor.org/stable/2987588}{The
  Statistician}}, volume~32, 1983.

\bibitem[\protect\citeauthoryear{Depeweg \bgroup \em et al.\egroup
  }{2018}]{depeweg2018decomposition}
Stefan Depeweg, Jose-Miguel Hernandez-Lobato, Finale Doshi-Velez, and Steffen
  Udluft.
\newblock {Decomposition of Uncertainty in Bayesian Deep Learning for Efficient
  and Risk-sensitive Learning}.
\newblock In {\em
  \href{https://proceedings.mlr.press/v80/depeweg18a.html}{JMLR}}, volume~80,
  pages 1184--1193, 2018.

\bibitem[\protect\citeauthoryear{Gawlikowski \bgroup \em et al.\egroup
  }{2021}]{gawlikowski2022survey}
Jakob Gawlikowski, Cedrique Rovile~Njieutcheu Tassi, Mohsin Ali, Jongseok Lee,
  Matthias Humt, Jianxiang Feng, Anna Kruspe, Rudolph Triebel, Peter Jung,
  Ribana Roscher, Muhammad Shahzad, Wen Yang, Richard Bamler, and Xiao~Xiang
  Zhu.
\newblock {A Survey of Uncertainty in Deep Neural Networks}.
\newblock In {\em \href{https://arxiv.org/abs/2107.03342}{arXiv:2107.03342}},
  July 2021.

\bibitem[\protect\citeauthoryear{G\'{o}mez \bgroup \em et al.\egroup
  }{2021}]{gomez_guan_tripathy}
Jairo~Alejandro G\'{o}mez, ChengHe Guan, Pratyush Tripathy, Juan~Carlos Duque,
  Santiago Passos, Michael Keith, and Jialin Liu.
\newblock {Analyzing the Spatiotemporal Uncertainty in Urbanization
  Predictions}.
\newblock In {\em \href{https://www.mdpi.com/2072-4292/13/3/512}{Remote
  Sensing}}, volume {13(512)}, 2021.

\bibitem[\protect\citeauthoryear{Guo \bgroup \em et al.\egroup
  }{2017}]{guo2017calibration}
Chuan Guo, Geoff Pleiss, Yu~Sun, and Kilian~Q. Weinberger.
\newblock {On Calibration of Modern Neural Networks}.
\newblock In {\em \href{https://dl.acm.org/doi/10.5555/3305381.3305518}{ICML}},
  volume~70, pages 1321--1330, August 2017.

\bibitem[\protect\citeauthoryear{Hoeting \bgroup \em et al.\egroup
  }{1999}]{hoeting1999bayesian}
Jennifer~A. Hoeting, David Madigan, Adrian~E. Raftery, and Chris~T. Volinsky.
\newblock {Bayesian Model Averaging: A Tutorial}.
\newblock In {\em
  \href{https://projecteuclid.org/journals/statistical-science/volume-14/issue-4/Bayesian-model-averaging--a-tutorial-with-comments-by-M/10.1214/ss/1009212519.full}{Statist.
  Sci.}}, volume {14(4)}, pages 382--417, November 1999.

\bibitem[\protect\citeauthoryear{Hollemans}{2020}]{hollance2020rel}
Matthijs Hollemans.
\newblock {Reliability Diagrams}.
\newblock \url{https://github.com/hollance/reliability-diagrams}, 2020.

\bibitem[\protect\citeauthoryear{Hüllermeier and
  Waegeman}{2021}]{huellermeier_waegeman}
Eyke Hüllermeier and Willem Waegeman.
\newblock {Aleatoric and Epistemic Uncertainty in Machine Learning: An
  Introduction to Concepts and Methods}.
\newblock In {\em
  \href{https://link.springer.com/article/10.1007/s10994-021-05946-3}{Machine
  Learning}}, volume 110, pages 457--506, March 2021.

\bibitem[\protect\citeauthoryear{Izmailov \bgroup \em et al.\egroup
  }{2018}]{izmailov2018averaging}
Pavel Izmailov, Dmitrii Podoprikhin, Timur Garipov, Dmitry Vetrov, and
  Andrew~Gordon Wilson.
\newblock {Averaging Weights Leads to Wider Optima and Better Generalization}.
\newblock In {\em
  \href{http://auai.org/uai2018/proceedings/papers/313.pdf}{UAI}}, 2018.

\bibitem[\protect\citeauthoryear{Kendall and
  Gal}{2017}]{kendall2017uncertainties}
Alex Kendall and Yarin Gal.
\newblock {What Uncertainties Do We Need in Bayesian Deep Learning for Computer
  Vision}.
\newblock In {\em \href{https://dl.acm.org/doi/10.5555/3295222.3295309}{NIPS}},
  volume~30, pages 5580--5590, December 2017.

\bibitem[\protect\citeauthoryear{Khan \bgroup \em et al.\egroup
  }{2018}]{khan2018fast}
Mohammad Khan, Didrik Nielsen, Voot Tangkaratt, Wu~Lin, Yarin Gal, and Akash
  Srivastava.
\newblock {Fast and Scalable Bayesian Deep Learning by Weight-Perturbation in
  Adam}.
\newblock In {\em \href{https://proceedings.mlr.press/v80/khan18a.html}{JMLR}},
  volume~80, pages 2611--2620, 2018.

\bibitem[\protect\citeauthoryear{Kwon \bgroup \em et al.\egroup
  }{2018}]{kwon2018uncertainty}
Yongchan Kwon, Joong-Ho Won, Beom~Joan Kim, and Myunghee~Cho Paik.
\newblock {Uncertainty Quantification Using Bayesian Neural Networks in
  Classification: Application to Ischemic Stroke Lesion Segmentation}.
\newblock In {\em \href{https://openreview.net/forum?id=Sk_P2Q9sG}{ICLR}},
  2018.

\bibitem[\protect\citeauthoryear{Lakshminarayanan \bgroup \em et al.\egroup
  }{2017}]{lakshminarayanan2017simple}
Balaji Lakshminarayanan, Ale{-}xander Pritzel, and Charles Blundell.
\newblock {Simple and Scalable Predictive Uncertainty Estimation Using Deep
  Ensembles}.
\newblock In {\em \href{https://dl.acm.org/doi/10.5555/3295222.3295387}{NIPS}},
  pages 6405--6416, December 2017.

\bibitem[\protect\citeauthoryear{Li \bgroup \em et al.\egroup
  }{2022}]{li2022uncertainty}
Xiaotong Li, Yongxing Dai, Yixiao Ge, Jun Liu, Ying Shan, and Lingyu Duan.
\newblock {Uncertainty Modeling for Out-of-Distribution Generalization}.
\newblock In {\em \href{https://openreview.net/pdf?id=6HN7LHyzGgC}{ICLR}},
  2022.

\bibitem[\protect\citeauthoryear{Long \bgroup \em et al.\egroup
  }{2014}]{long_wang}
Mingsheng Long, Jianmin Wang, Guiguang Ding, Jiaguang Sun, and Philip~S. Yu.
\newblock {Transfer Joint Matching for Unsupervised Domain Adaptation}.
\newblock In {\em \href{https://ieeexplore.ieee.org/document/6909579}{CVPR}},
  pages 1410--1417, Columbus, OH, June 2014.

\bibitem[\protect\citeauthoryear{Maddox \bgroup \em et al.\egroup
  }{2019}]{maddox2019simple}
Wesley~J. Maddox, Timur Garipov, Pavel Izmailov, Dmitry Vetrov, and
  Andrew~Gordon Wilson.
\newblock {A Simple Baseline for Bayesian Uncertainty in Deep Learning}.
\newblock In {\em \href{https://dl.acm.org/doi/10.5555/3454287.3455466}{NIPS}},
  pages 13153--13164, December 2019.

\bibitem[\protect\citeauthoryear{Ott \bgroup \em et al.\egroup }{2020}]{ott}
Felix Ott, Mohamad Wehbi, Tim Hamann, Jens Barth, Björn Eskofier, and
  Christopher Mutschler.
\newblock {The OnHW Dataset: Online Handwriting Recognition from IMU-Enhanced
  Ballpoint Pens with Machine Learning}.
\newblock In {\em \href{https://doi.org/10.1145/3411842}{IMWUT}}, volume {4(3),
  article 92}, Canc\'{u}n, Mexico, September 2020.

\bibitem[\protect\citeauthoryear{Ott \bgroup \em et al.\egroup
  }{2022a}]{ott_ijcai}
Felix Ott, David Rügamer, Lucas Heublein, Bernd Bischl, and Christopher
  Mutschler.
\newblock {Cross-Modal Common Representation Learning with Triplet Loss
  Functions}.
\newblock In {\em \href{https://arxiv.org/abs/2202.07901}{arXiv:2202.07901}},
  February 2022.

\bibitem[\protect\citeauthoryear{Ott \bgroup \em et al.\egroup
  }{2022b}]{ott_mm}
Felix Ott, David Rügamer, Lucas Heublein, Bernd Bischl, and Christopher
  Mutschler.
\newblock {Domain Adaptation for Time-Series Classification to Mitigate
  Covariate Shift}.
\newblock In {\em \href{https://arxiv.org/abs/2204.03342}{arXiv:2204.03342}},
  April 2022.

\bibitem[\protect\citeauthoryear{Ott \bgroup \em et al.\egroup
  }{2022c}]{ott_wacv}
Felix Ott, David Rügamer, Lucas Heublein, Bernd Bischl, and Christopher
  Mutschler.
\newblock {Joint Classification and Trajectory Regression of Online Handwriting
  using a Multi-Task Learning Approach}.
\newblock In {\em
  \href{https://openaccess.thecvf.com/content/WACV2022/html/Ott_Joint_Classification_and_Trajectory_Regression_of_Online_Handwriting_Using_a_WACV_2022_paper.html}{WACV}},
  pages 266--276, Waikoloa, HI, January 2022.

\bibitem[\protect\citeauthoryear{Ott \bgroup \em et al.\egroup
  }{2022d}]{ott_ijdar}
Felix Ott, David Rügamer, Lucas Heublein, Tim Hamann, Jens Barth, Bernd
  Bischl, and Christopher Mutschler.
\newblock {Benchmarking Online Sequence-to-Sequence and Character-based
  Handwriting Recognition from IMU-Enhanced Pens}.
\newblock In {\em \href{https://arxiv.org/abs/2202.07036}{arXiv:2202.07036}},
  February 2022.

\bibitem[\protect\citeauthoryear{Ovadia \bgroup \em et al.\egroup
  }{2019}]{ovadia2019can}
Yaniv Ovadia, Emily Fertig, Jie Ren, Zachary Nado, D.~Sculley, Sebastian
  Nowozin, Joshua~V. Dillon, Balaji Lakshminarayanan, and Jasper Snoek.
\newblock {Can You Trust Your Model's Uncertainty? Evaluating Predictive
  Uncertainty Under Dataset Shift}.
\newblock In {\em
  \href{https://dl.acm.org/doi/abs/10.5555/3454287.3455541}{NIPS}}, volume~32,
  pages 14003--14014, December 2019.

\bibitem[\protect\citeauthoryear{Pan and Yang}{2009}]{pan_yang}
Sinno~Jialin Pan and Qiang Yang.
\newblock {A Survey on Transfer Learning}.
\newblock In {\em \href{https://ieeexplore.ieee.org/document/5288526}{Trans. on
  Knowledge and Data Engineering}}, volume {22(10)}, pages 1345--1359, October
  2009.

\bibitem[\protect\citeauthoryear{Plamondon and Srihari}{2000}]{plamondon}
Rejean Plamondon and Sargur~N. Srihari.
\newblock {On-line and Off-line Handwriting Recognition: A Comprehensive
  Survey}.
\newblock In {\em \href{https://ieeexplore.ieee.org/document/824821}{TPAMI}},
  volume 22(1), pages 63--84, January 2000.

\bibitem[\protect\citeauthoryear{Saenko \bgroup \em et al.\egroup
  }{2010}]{saenko}
Kate Saenko, Brian Kulis, Mario Fritz, and Trevor Darrell.
\newblock {Adapting Visual Category Models to New Domains}.
\newblock In {\em
  \href{https://link.springer.com/chapter/10.1007/978-3-642-15561-1_16}{ECCV}},
  volume 6314, pages 213--226, 2010.

\bibitem[\protect\citeauthoryear{Schölkopf \bgroup \em et al.\egroup
  }{2021}]{schoelkopf_locatello}
Bernhard Schölkopf, Francesco Locatello, Stefan Bauer, Nan~Rosemary Ke, Nal
  Kalchbrenner, Anirudh Goyal, and Yoshua Bengio.
\newblock {Toward Causal Representation Learning}.
\newblock In {\em
  \href{https://cardiacmr.hms.harvard.edu/files/cardiacmr/files/toward_causal_representation_learning.pdf}{Proceedings
  of the IEEE}}, volume {109(5)}, pages 612--634, 2021.

\bibitem[\protect\citeauthoryear{Shao \bgroup \em et al.\egroup
  }{2014}]{shao_zhu}
Ling Shao, Fan Zhu, and Xuelong Li.
\newblock {Transfer Learning for Visual Categorization: A Survey}.
\newblock In {\em \href{https://ieeexplore.ieee.org/document/6847217}{Trans. on
  Neural Networks and Learning Systems}}, volume {26(5)}, pages 1019--1034,
  July 2014.

\bibitem[\protect\citeauthoryear{Smith and Gal}{2018}]{smith2018understanding}
Lewis Smith and Yarin Gal.
\newblock {Understanding Measures of Uncertainty for Adversarial Example
  Detection}.
\newblock In {\em
  \href{http://auai.org/uai2018/proceedings/papers/207.pdf}{UAI}}, 2018.

\bibitem[\protect\citeauthoryear{Sun \bgroup \em et al.\egroup
  }{2016}]{sun_feng}
Baochen Sun, Jiashin Feng, and Kate Saenko.
\newblock {Correlation Alignment for Unsupervised Domain Adaptation}.
\newblock In {\em
  \href{https://arxiv.org/pdf/1612.01939.pdf}{arXiv:1612.01939}}, December
  2016.

\bibitem[\protect\citeauthoryear{Wu \bgroup \em et al.\egroup
  }{2021}]{wu_gao_xiong}
Dongxia Wu, Liyao Gao, Xinyue Xiong, Matteo Chinazzi, Alessandro Vespignani,
  Yi-An Ma, and Rose Yu.
\newblock {Quantifying Uncertainty in Deep Spatiotemporal Forecasting}.
\newblock In {\em \href{https://arxiv.org/abs/2105.11982}{arXiv:2105.11982}},
  May 2021.

\bibitem[\protect\citeauthoryear{Zhou \bgroup \em et al.\egroup
  }{2021}]{zhou_wang_xie}
Zhengyang Zhou, Yang Wang, Xike Xie, Lei Qiao, and Yuantao Li.
\newblock {STUaNet: Understanding Uncertainty in Spatiotemporal Collective
  Human Mobility}.
\newblock In {\em \href{https://dl.acm.org/doi/10.1145/3442381.3449817}{WWW}},
  pages 1868--1879, April 2021.

\end{thebibliography}

\normalsize

\begin{figure*}[t!]
\centering
\begin{subfigure}{\subfigsizeb\textwidth}
    \centering
    \includegraphics[width=1\linewidth]{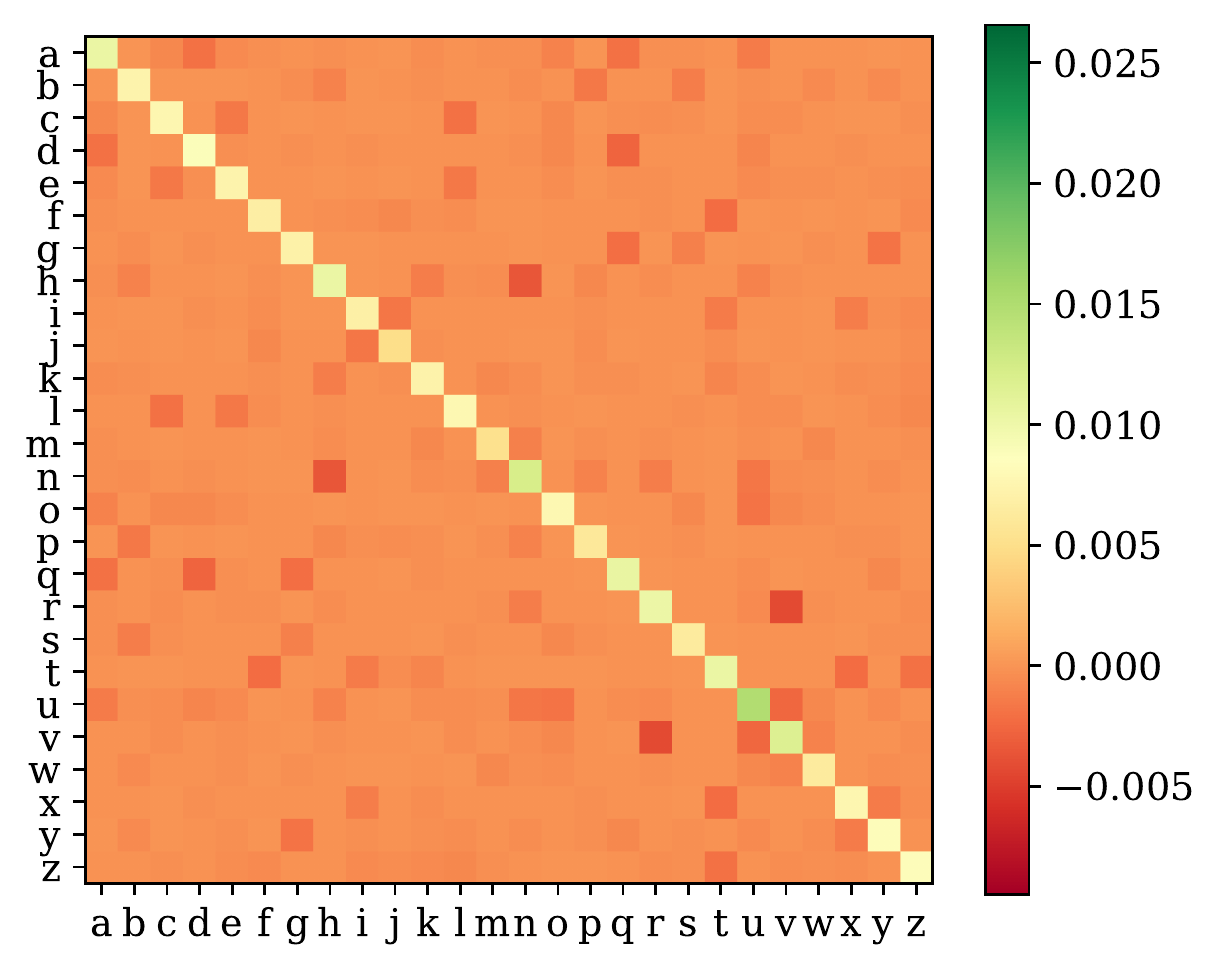}
    \vspace{-0.4cm}
    \caption{Aleatoric uncertainty.}
    \label{fig_app_alea_low}
\end{subfigure}
\hfill
\begin{subfigure}{\subfigsizeb\textwidth}
    \centering
    \includegraphics[width=1\linewidth]{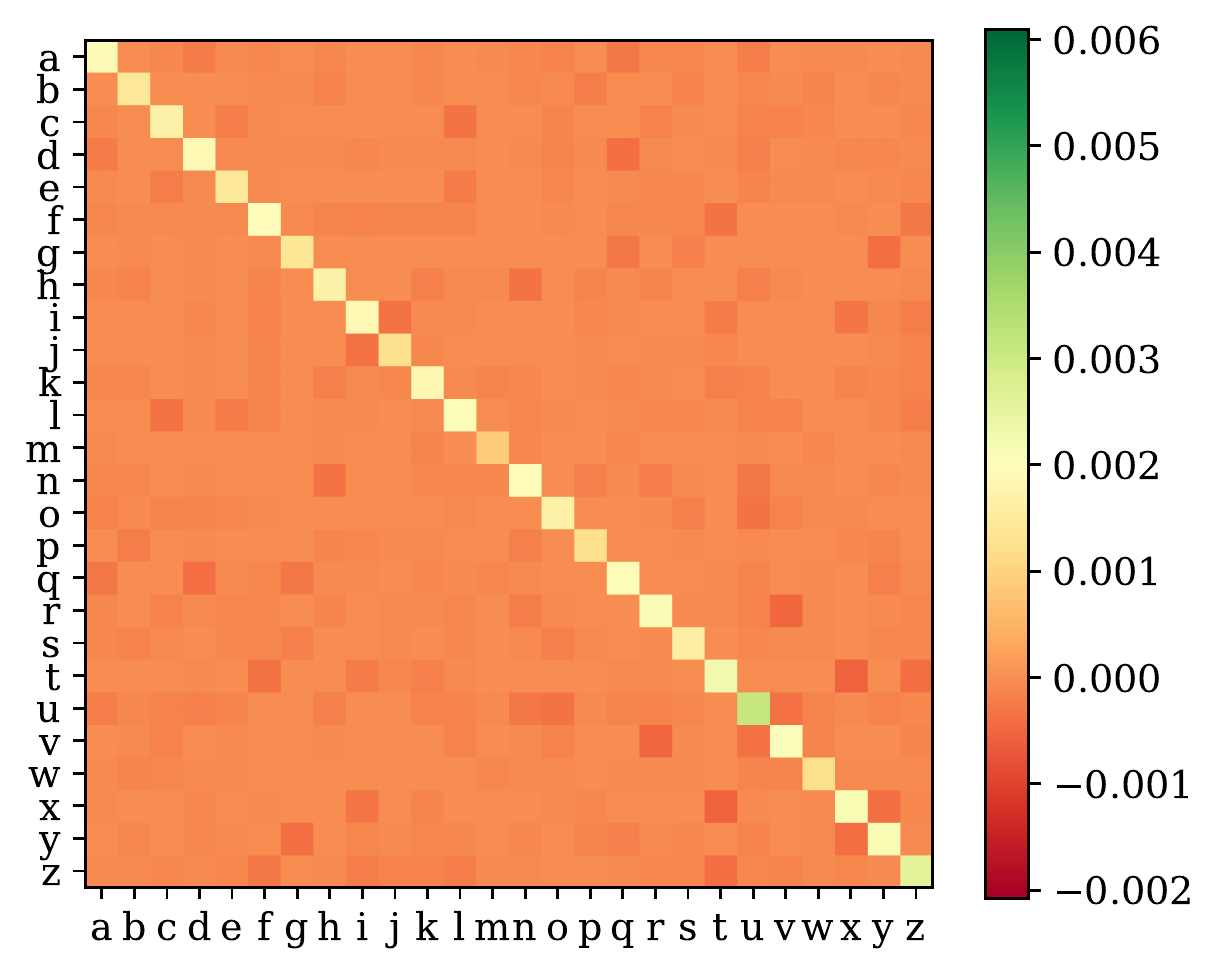} 
    \vspace{-0.4cm}
    \caption{Epistemic uncertainty.}
    \label{fig_app_epis_low}
\end{subfigure}
\hfill
\begin{subfigure}{\subfigsizeb\textwidth}
    \centering
    \includegraphics[width=1\linewidth]{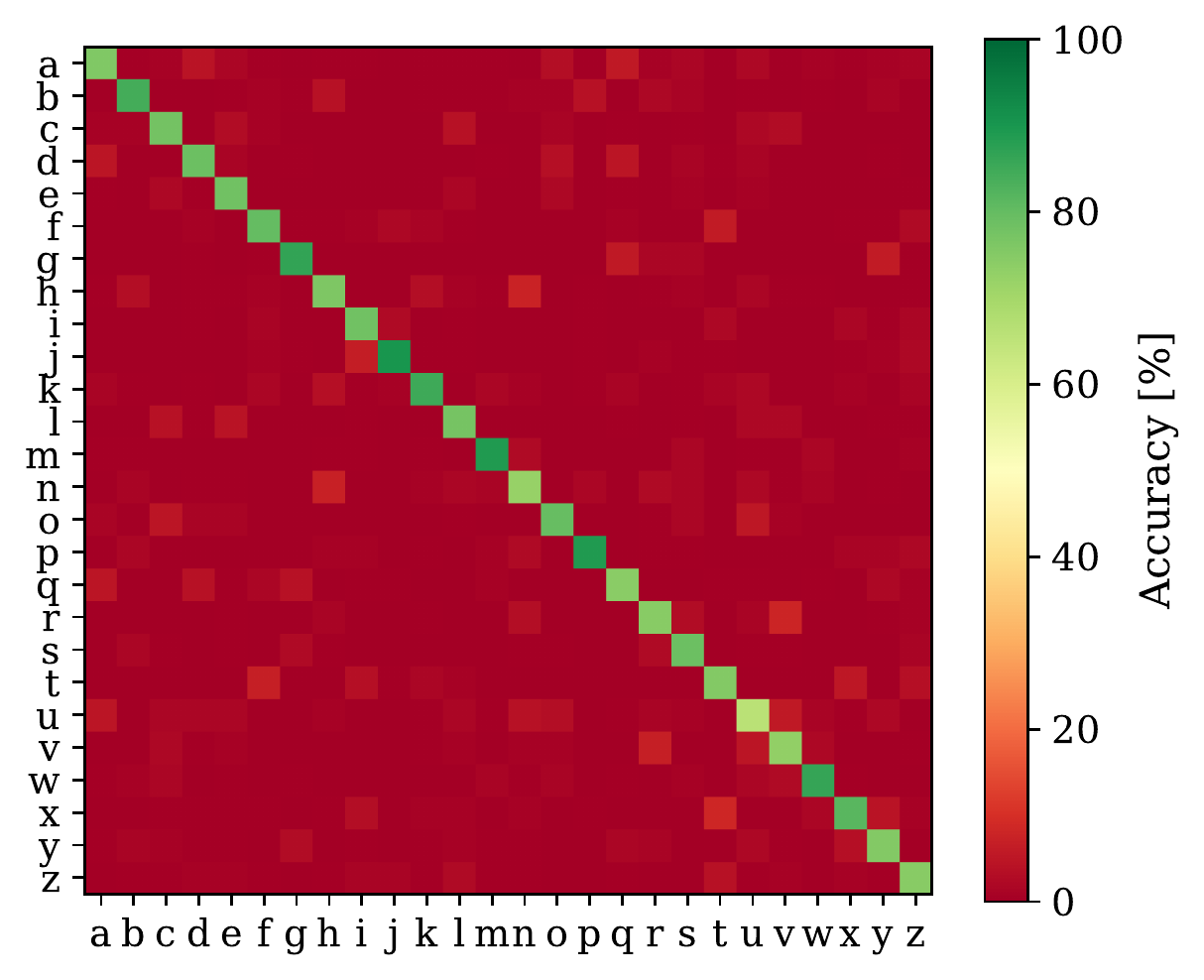} 
    \vspace{-0.4cm}
    \caption{Confusion matrix of accuracy.}
    \label{fig_app_conf_low}
\end{subfigure}
\vspace{-0.2cm}
\caption{Uncertainty prediction for the Deep Ensemble CNN+BiLSTM model (which outperformed the TCN-based architecture) trained on the lowercase WD (right-handed only) dataset. Note that the color scale is fixed for all subplots for comparability with Figure~\ref{fig:both} and \ref{fig:uppercase_conf}.}
\label{fig:lowercase_conf}
\end{figure*}

\appendix
\section{Appendices}
\label{sec_appendix}

We propose model parameters in Section~\ref{sec_app_parameters} and show an evaluation per character in Section~\ref{sec_app_character}. We propose results for the SWAG model in Section~\ref{app_swag_results}. 

\subsection{Model and UQ Method Parameters}
\label{sec_app_parameters}

For reproducibility, we state all general model architecture parameters and propose training parameters for the SWAG model. For all experiments we use Nvidia Tesla V100-SXM2 GPUs with 32 GB VRAM coupled with Intel Core Xeon CPUs and 192 GB RAM.

\paragraph{Model Parameters.} We use a CNN with dropout rate 20\%, convolutional layers with kernel size 4 and filter size 200. The temporal cell (LSTM, BiLSTM or TCN) contains 100, 100 or 120 neurons, respectively. We interpolate the time-series to 64 time steps, and train the model for 2,000 epochs with early stopping and a batch size of 50.

\paragraph{SWAG Parameters.} We initialize the stochastic gradient descent (SGD) optimizer with initial learning rate $10^{-2}$, a momentum of 0.9, and weight decay of $10^{-4}$. The stochastic weight averaging (SWA) burn-in period was run for 10 epochs. SWAG showed a training process with fast convergence.



\subsection{Evaluation per Character}
\label{sec_app_character}

\paragraph{Confusion Matrices.} We propose the confusion matrices for the aleatoric and epistemic uncertainty as well as the accuracy (in \%) for the uppercase (see Figure~\ref{fig:uppercase_conf}) and lowercase (see Figure~\ref{fig:lowercase_conf}) datasets. While for the combined training, lower- and uppercase characters are often misclassified, the separate training leads to confusion of characters with similar shapes, e.g., for the uppercase task, the model is uncertain for \texttt{"D"} and \texttt{"P"}, \texttt{"U"} and \texttt{"V"}, and \texttt{"T"} and \texttt{"X"}. These confusions can be identified with the aleatoric and epistemic uncertainty and correspond with the classification accuracies. Overall, the uncertainty for lowercase characters is higher (see Figure~\ref{fig_app_alea_low}) since the writing style of lowercase characters is oftentimes quite similar, e.g., \texttt{"r"} and \texttt{"v"}, \texttt{"u"} and \texttt{"v"}, \texttt{"h"} and \texttt{"n"}, and \texttt{"d"} and \texttt{"q"}. This also leads to a lower classification accuracy (see Figure~\ref{fig_app_conf_low}).

\begin{figure}[t!]
	\centering
	\begin{minipage}[b]{0.51\linewidth}
        \centering
        \includegraphics[trim=0 8 0 6, clip, width=1.0\columnwidth]{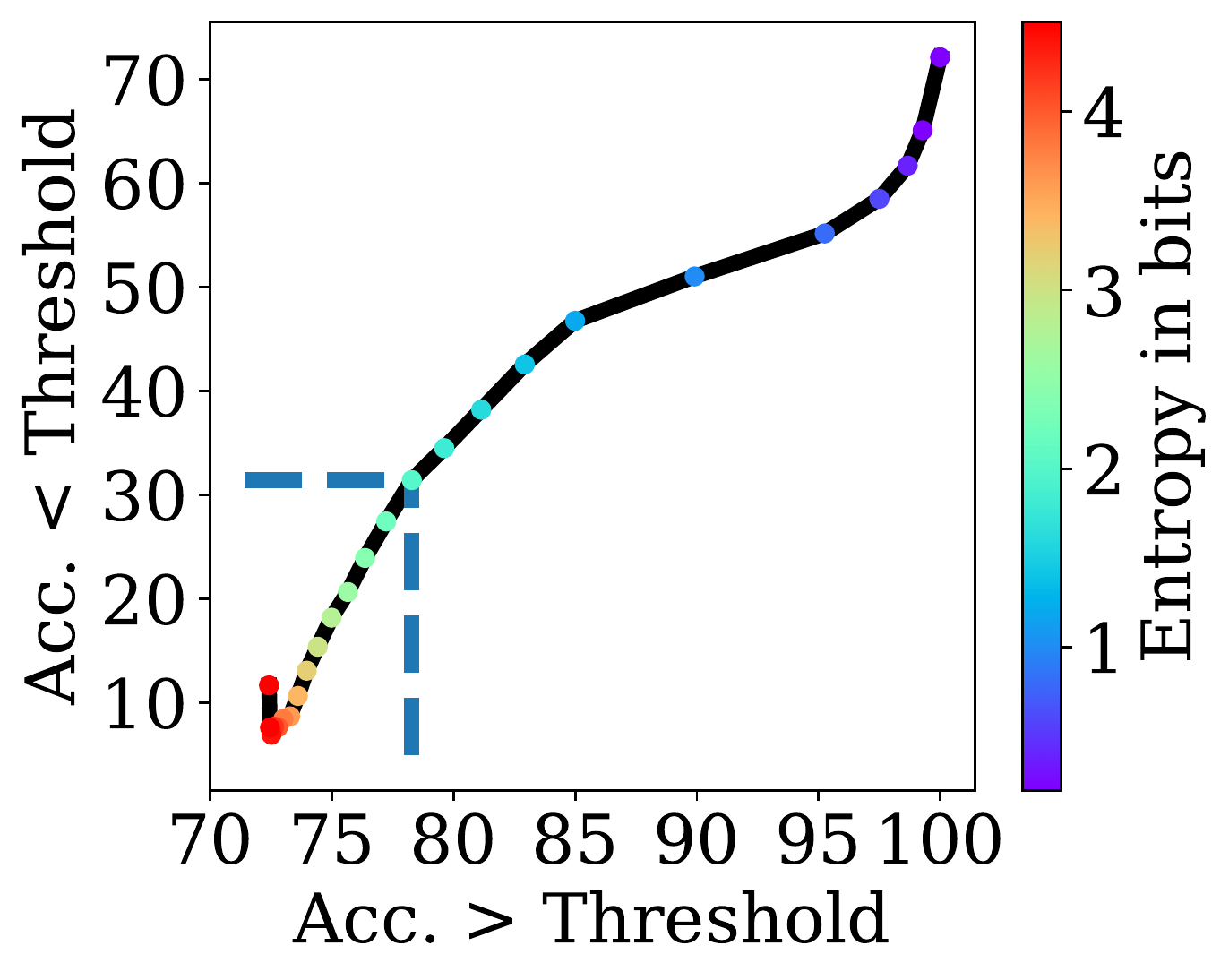}
        \subcaption{Sample accuracies below and above an entropy threshold.}
        \label{fig:trade-off_app}
    \end{minipage}
    \hfill
	\begin{minipage}[b]{0.47\linewidth}
        \centering
        \includegraphics[trim=0 8 0 6, clip, width=1.0\columnwidth]{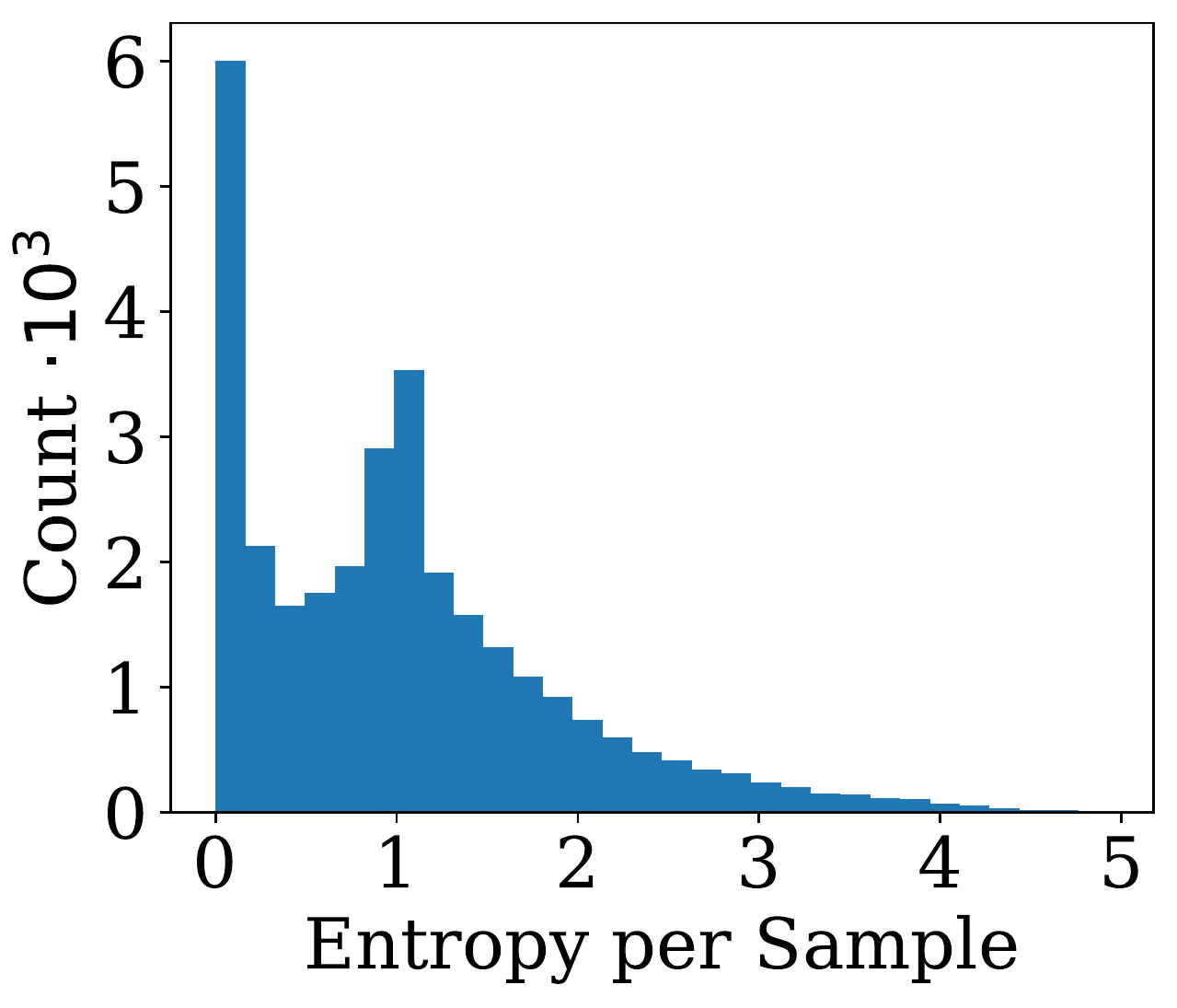}
        \subcaption{Histogram visualizing the entropy distribution.}
        \label{fig:hist_en_app}
    \end{minipage}
    \caption{Accuracy and entropy for the SWAG CNN+TCN model trained on the combined WD (right-handed only) dataset.}
    \label{fig:cutoff_appendix}
\end{figure}

\paragraph{Mutual Information and Entropy.} Figure~\ref{fig:hist_mi} shows the mutual information (MI) per character, and Figure~\ref{fig:hist_entropy} shows the entropy, respectively. In general, the MI and entropy correlates and are similar for each character. The MI and entropy is high for the characters \texttt{"U"}, \texttt{"u"}, \texttt{"v"}, \texttt{"x"}, and \texttt{"z"}. Furthermore, both metrics are higher for lowercase characters than for uppercase characters. This corresponds to the confusion matrices in Figure~\ref{fig:uppercase_conf} and ~\ref{fig:lowercase_conf} where aleatoric uncertainty is higher for off-diagonals for lowercase characters.

\newcommand\subfigsizec{0.49}
\begin{figure*}[t!]
	\centering
	\begin{minipage}[b]{\subfigsizec\linewidth}
        \centering
        \includegraphics[trim=0 8 0 6, clip, width=1.0\columnwidth]{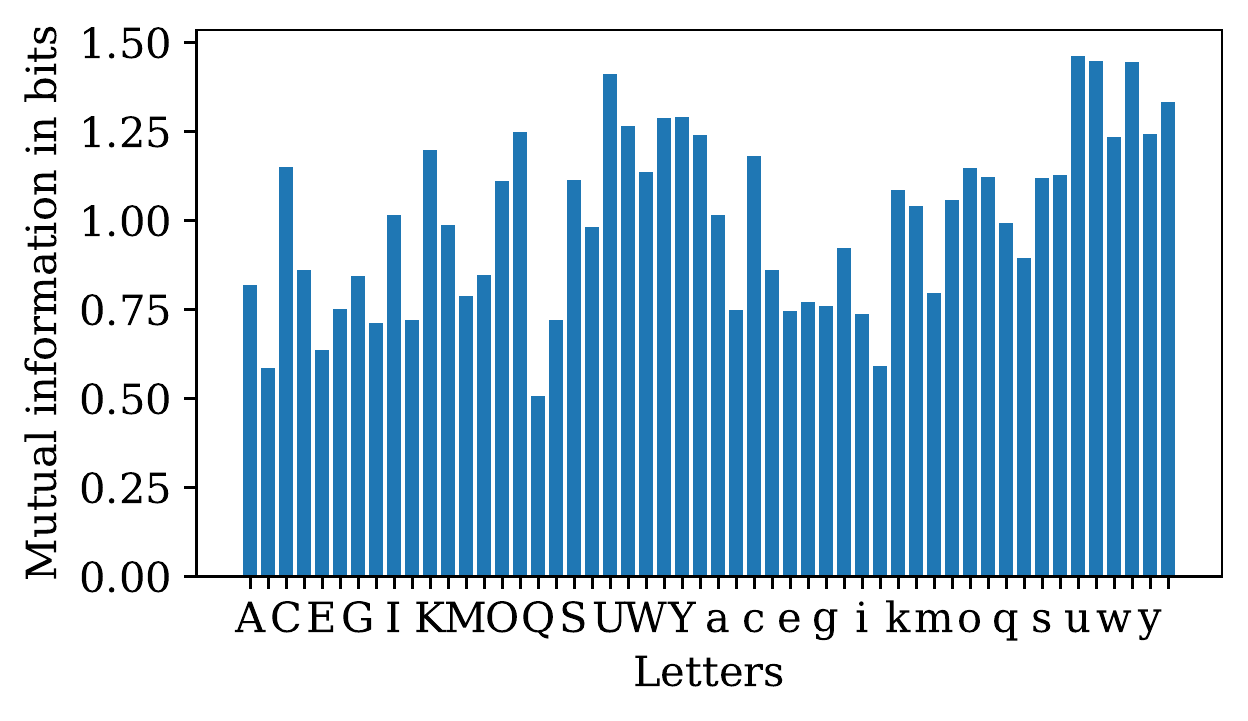}
        \subcaption{Mutual information per letter.}
        \label{fig:hist_mi}
    \end{minipage}
    \hfill
	\begin{minipage}[b]{\subfigsizec\linewidth}
        \centering
        \includegraphics[trim=0 10 0 6, clip, width=1.0\columnwidth]{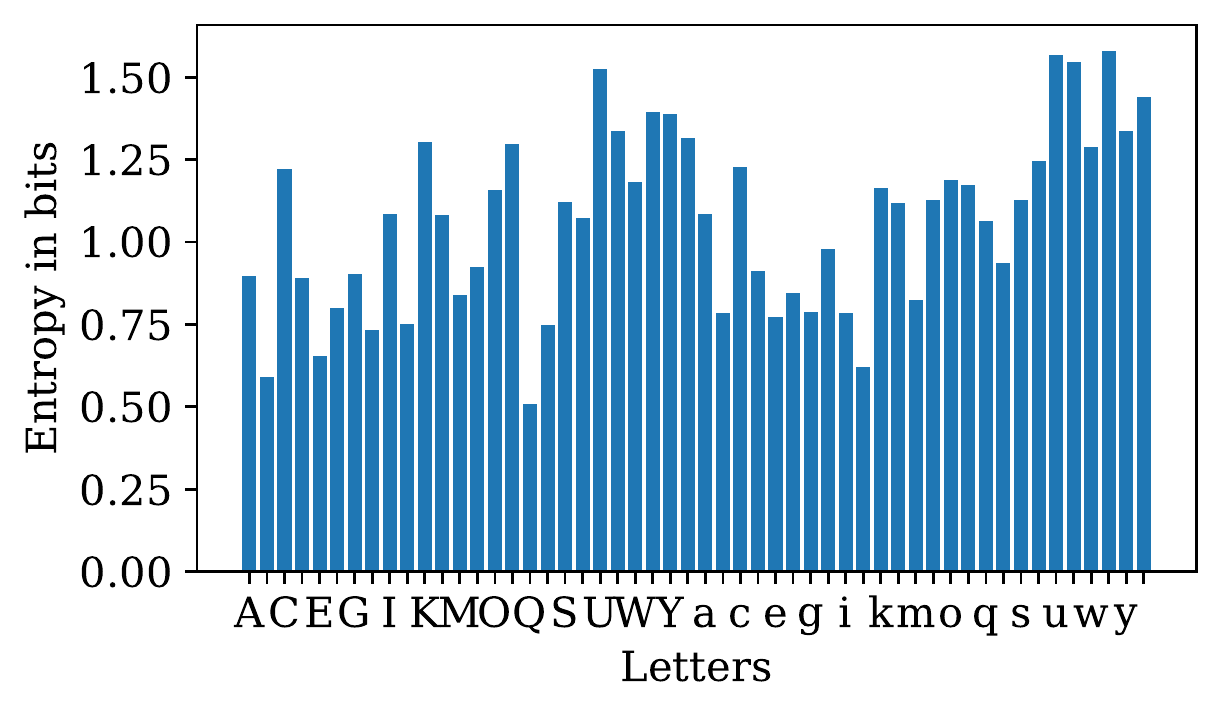}
        \subcaption{Entropy per letter.}
        \label{fig:hist_entropy}
    \end{minipage}
    \caption{Mutual information and entropy per letter for the Deep Ensemble CNN+TCN model trained on the combined WD (right-handed only) dataset. Note that we skiped every second character in the x-axis (ordered alphabetically) for readability.}
    \label{plot_mi_entropy_de}
\end{figure*}

\begin{figure*}[t!]
	\centering
	\begin{minipage}[b]{\subfigsizec\linewidth}
        \centering
        \includegraphics[trim=0 8 0 6, clip, width=1.0\columnwidth]{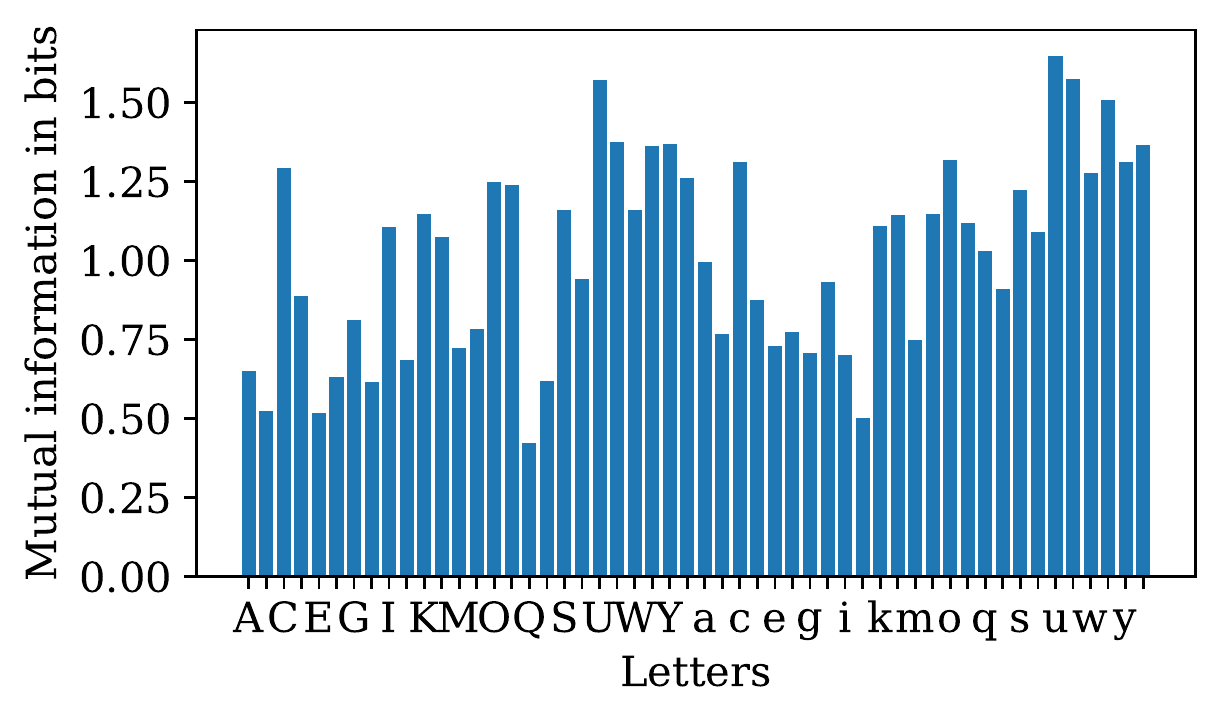}
        \subcaption{Mutual information per letter.}
        \label{fig:hist_mi_app}
    \end{minipage}
    \hfill
	\begin{minipage}[b]{\subfigsizec\linewidth}
        \centering
        \includegraphics[trim=0 10 0 6, clip, width=1.0\columnwidth]{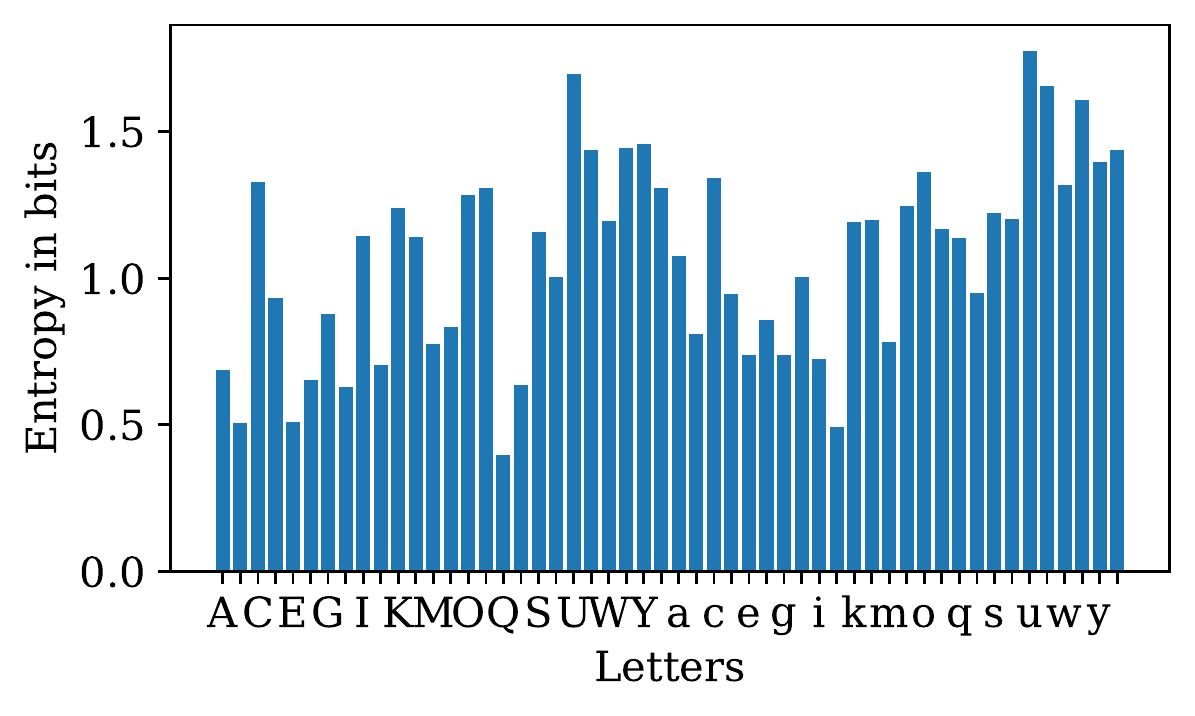}
        \subcaption{Entropy per letter.}
        \label{fig:hist_entropy_app}
    \end{minipage}
    \caption{Mutual information and entropy per letter for the SWAG CNN+TCN model trained on the combined WD (right-handed only) dataset. Note that we skipped every second character in the x-axis (ordered alphabetically) for readability.}
    \label{plot_mi_entropy_swag}
\end{figure*}

\subsection{SWAG Model Results}\label{app_swag_results}

\begin{figure*}[t!]
	\centering
	\begin{minipage}[b!]{\subfigsizea\linewidth}
        \centering
        \includegraphics[trim=5 7 7 7, clip, width=1.0\linewidth]{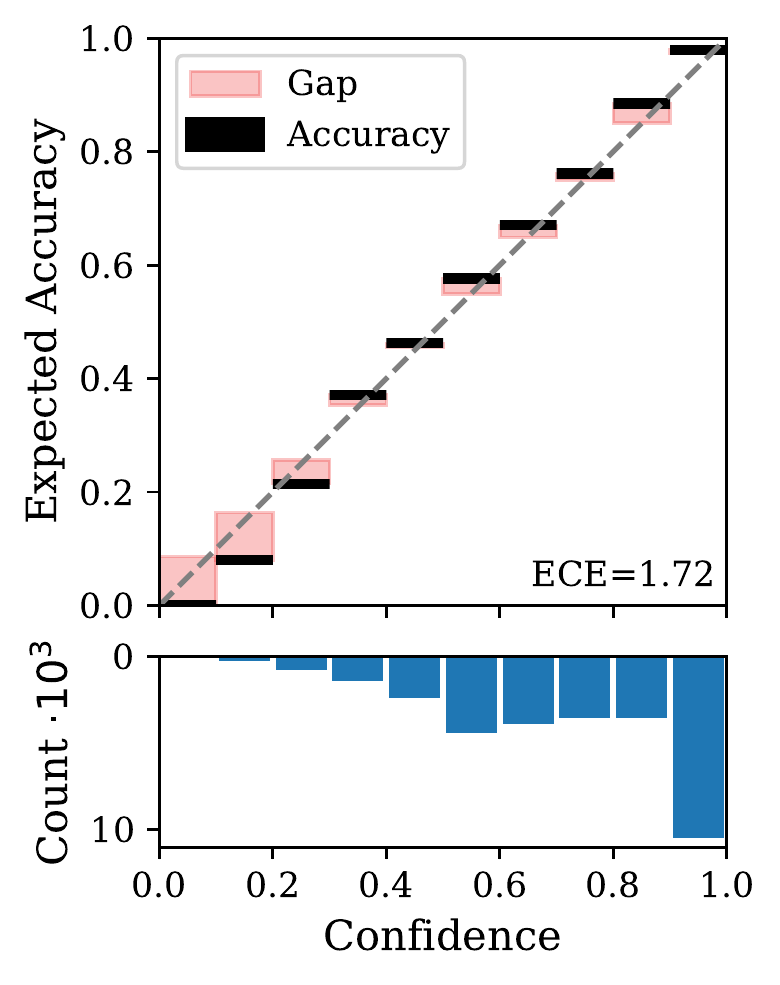}
        \subcaption{Evaluated on right-handed writers data.}
        \label{fig:eval_rel_dia_app_a}
    \end{minipage}
    \hfill
	\begin{minipage}[b!]{\subfigsizea\linewidth}
        \centering
        \includegraphics[trim=5 7 7 7, clip, width=1.0\linewidth]{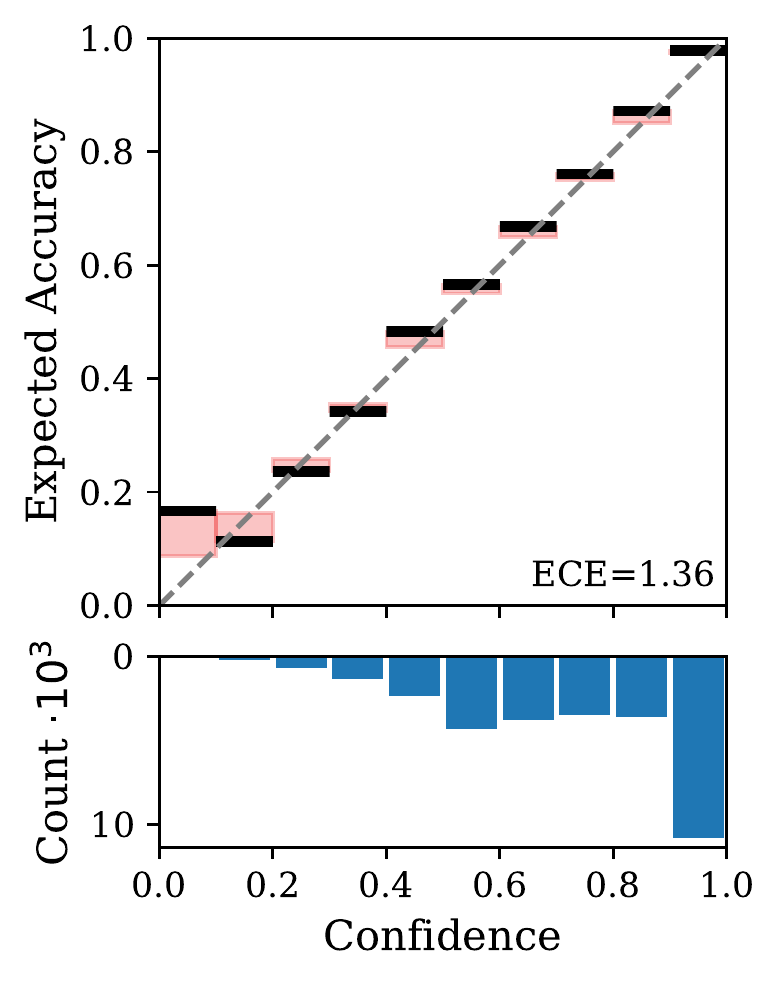}
        \subcaption{Evaluated on right-handed writers data.}
        \label{fig:eval_rel_dia_app_c}
    \end{minipage}
    \hfill
	\begin{minipage}[b!]{\subfigsizea\linewidth}
        \centering
        \includegraphics[trim=5 7 7 7, clip, width=1.0\linewidth]{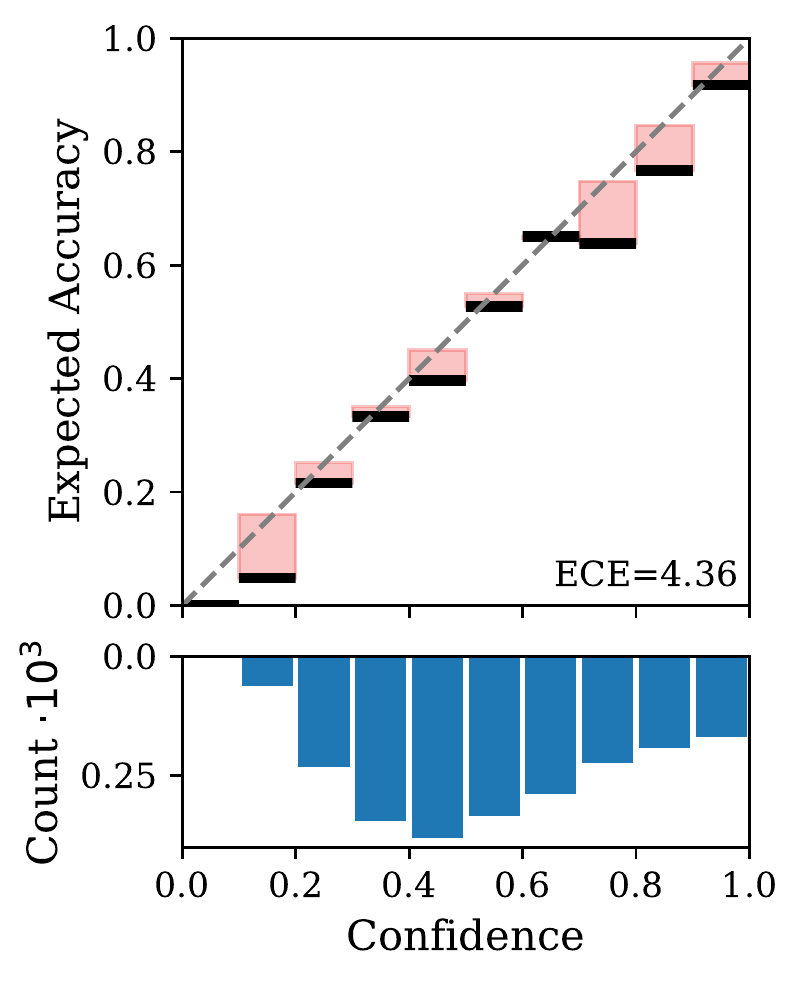}
        \subcaption{Evaluated on left-handed writers data.}
        \label{fig:eval_rel_dia_app_b}
    \end{minipage}
    \hfill
	\begin{minipage}[b!]{\subfigsizea\linewidth}
        \centering
        \includegraphics[trim=5 7 7 7, clip, width=1.0\linewidth]{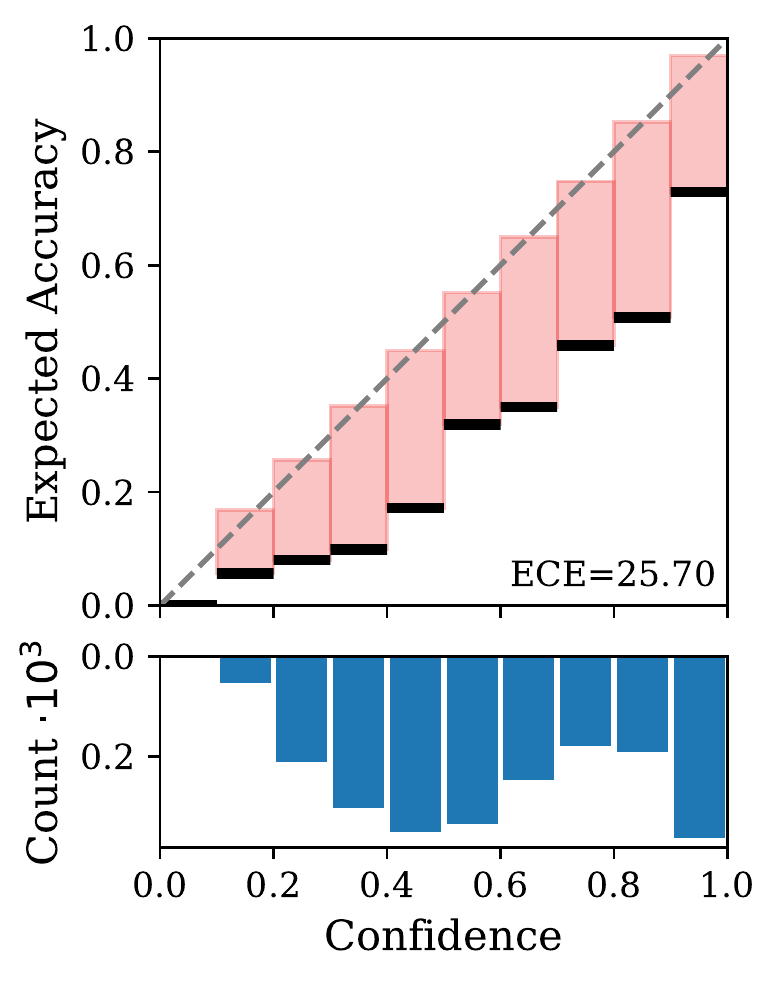}
        \subcaption{Evaluated on left-handed writers data.}
        \label{fig:eval_rel_dia_app_d}
    \end{minipage}
    \caption{Reliability diagram for the SWAG CNN+TCN model trained on the combined WD datasets. a) and c): Trained on the combined right- and left-handed writers datasets. b) and d): Trained on right-handed writers only.}
\label{fig:cal1_app}
\end{figure*}

\begin{figure*}[t!]
\centering
\begin{subfigure}{\subfigsizeb\textwidth}
    \centering
    \includegraphics[width=1.0\linewidth]{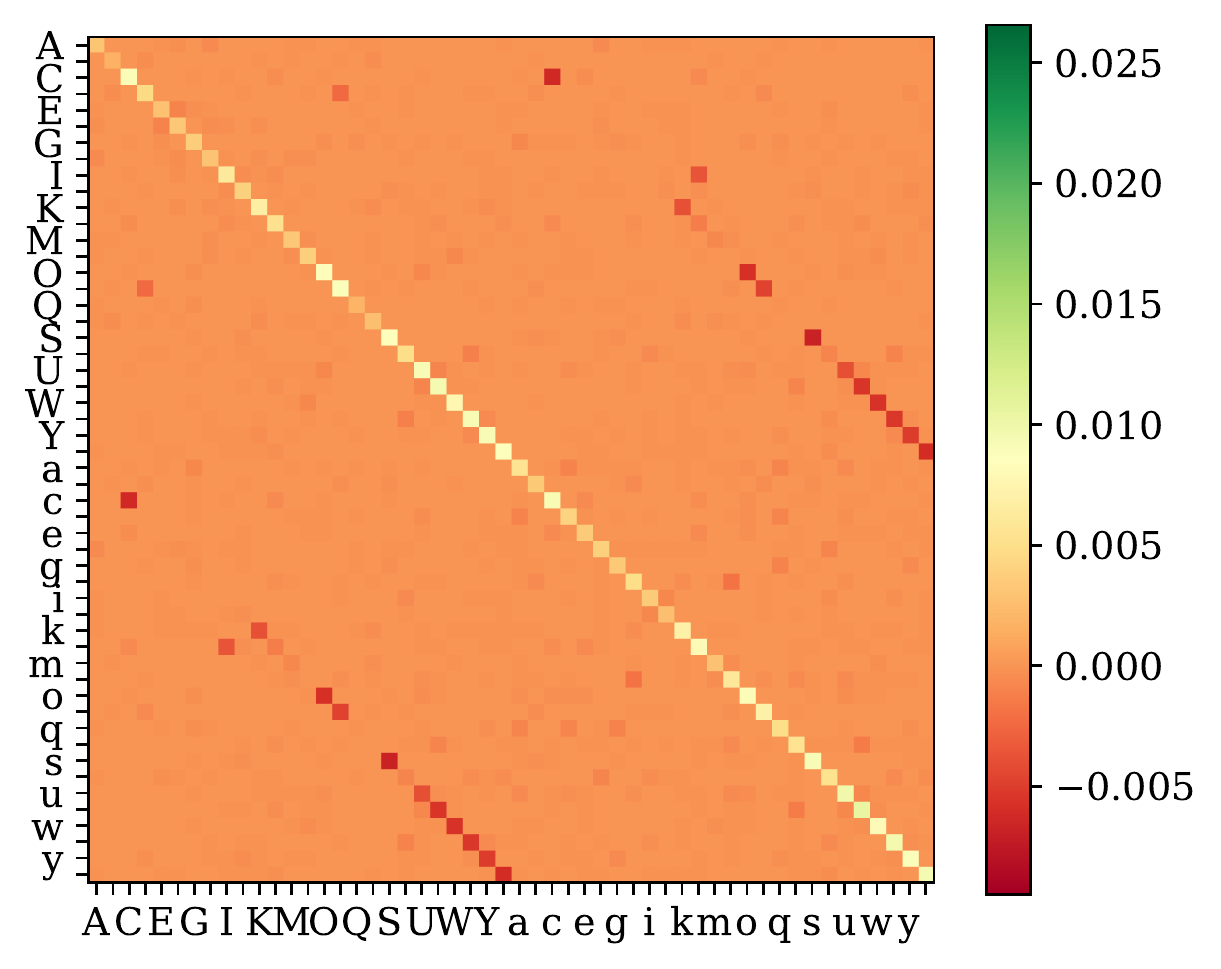}
    \vspace{-0.4cm}
    \caption{Aleatoric uncertainty.}
    \label{fig:alea_both_app}
\end{subfigure}
\hfill
\begin{subfigure}{\subfigsizeb\textwidth}
    \centering
    \includegraphics[width=1.0\linewidth]{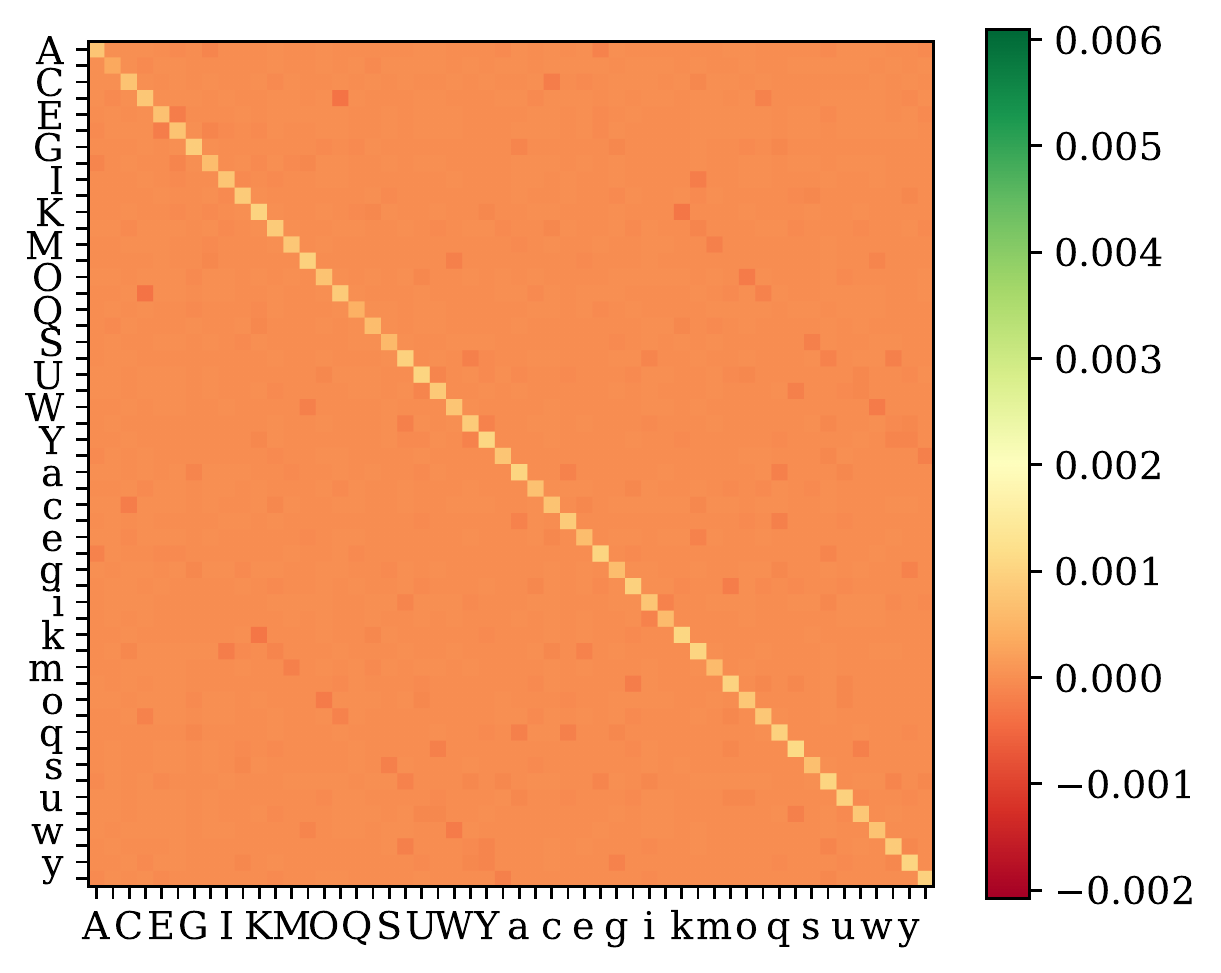} 
    \vspace{-0.4cm}
    \caption{Epistemic uncertainty.}
    \label{fig:epis_both_app}
\end{subfigure}
\hfill
\begin{subfigure}{\subfigsizeb\textwidth}
    \centering
    \includegraphics[width=1.0\linewidth]{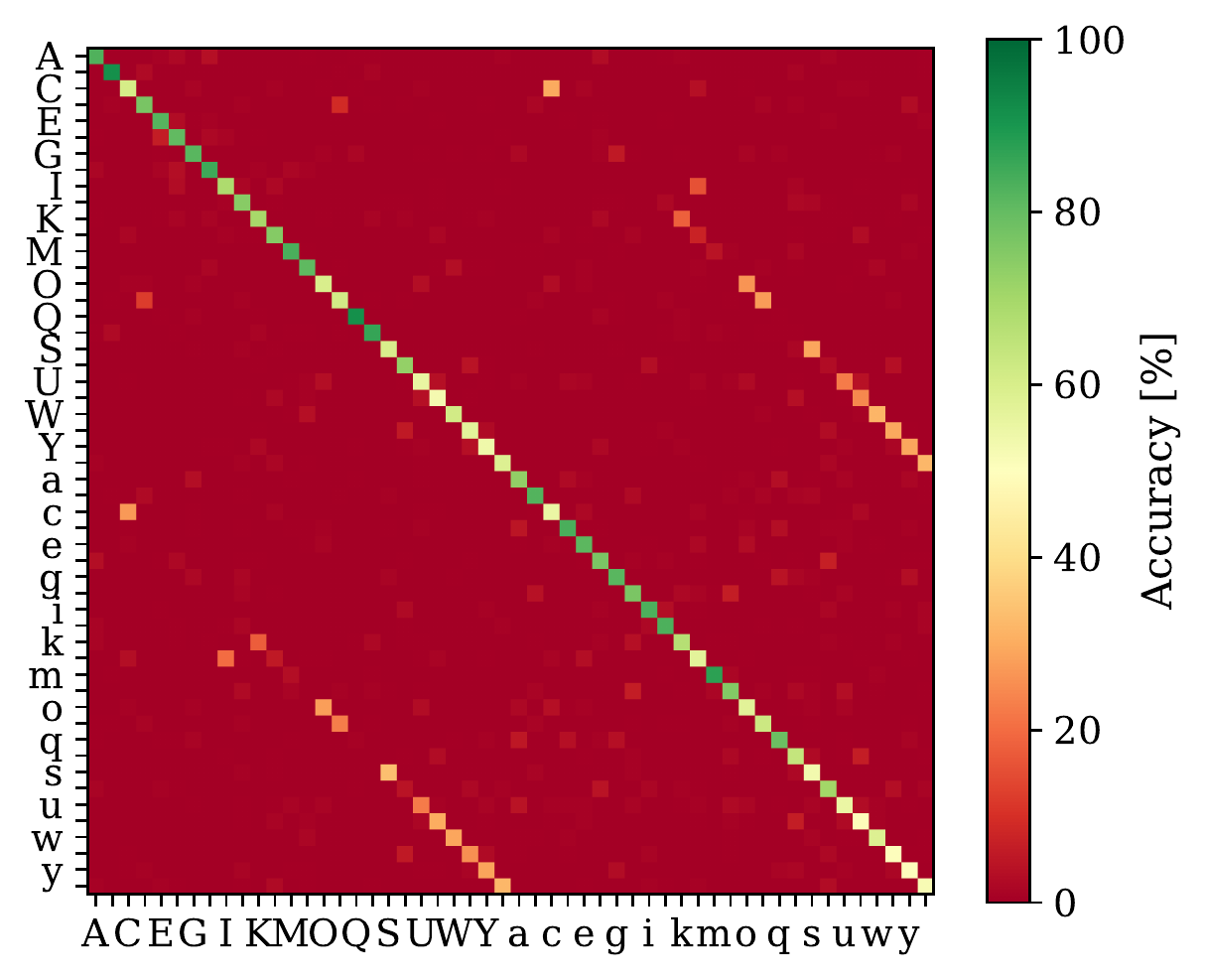} 
    \vspace{-0.4cm}
    \caption{Confusion matrix of accuracy.}
    \label{fig:conf_both_app}
\end{subfigure}
\caption{Uncertainty prediction for the SWAG CNN+TCN model trained on the combined WD (right-handed only) dataset. Note that the color scale is fixed for all subplots for comparability with the other heatmaps, and that we skipped every second character label for readability.}
\label{fig:both_app}
\end{figure*}

\begin{figure*}[t!]
\centering
\begin{subfigure}{\subfigsizeb\textwidth}
    \centering
    \includegraphics[width=1\linewidth]{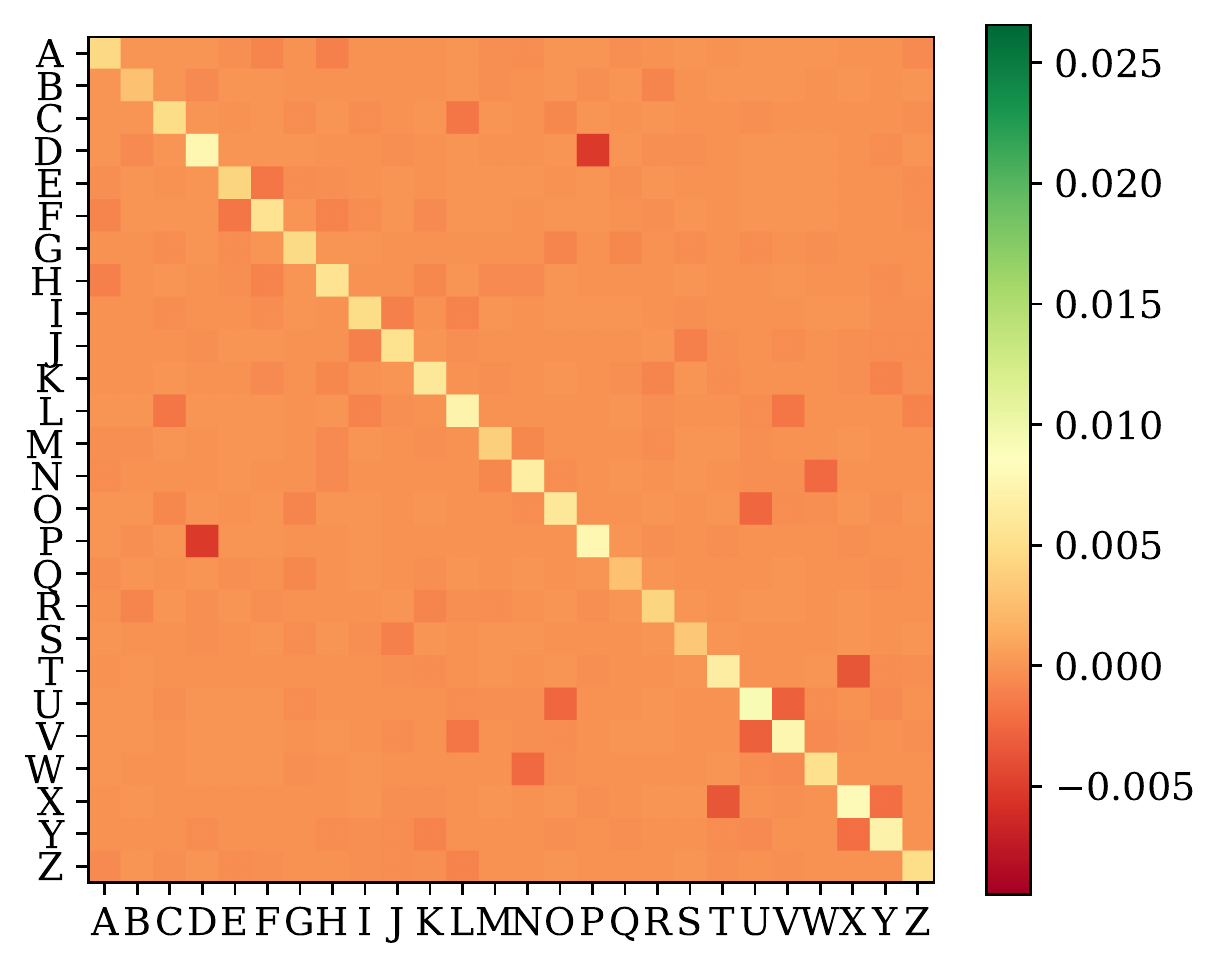}
    \vspace{-0.4cm}
    \caption{Aleatoric uncertainty.}
    \label{fig_app_alea_upp_app}
\end{subfigure}
\hfill
\begin{subfigure}{\subfigsizeb\textwidth}
    \centering
    \includegraphics[width=1\linewidth]{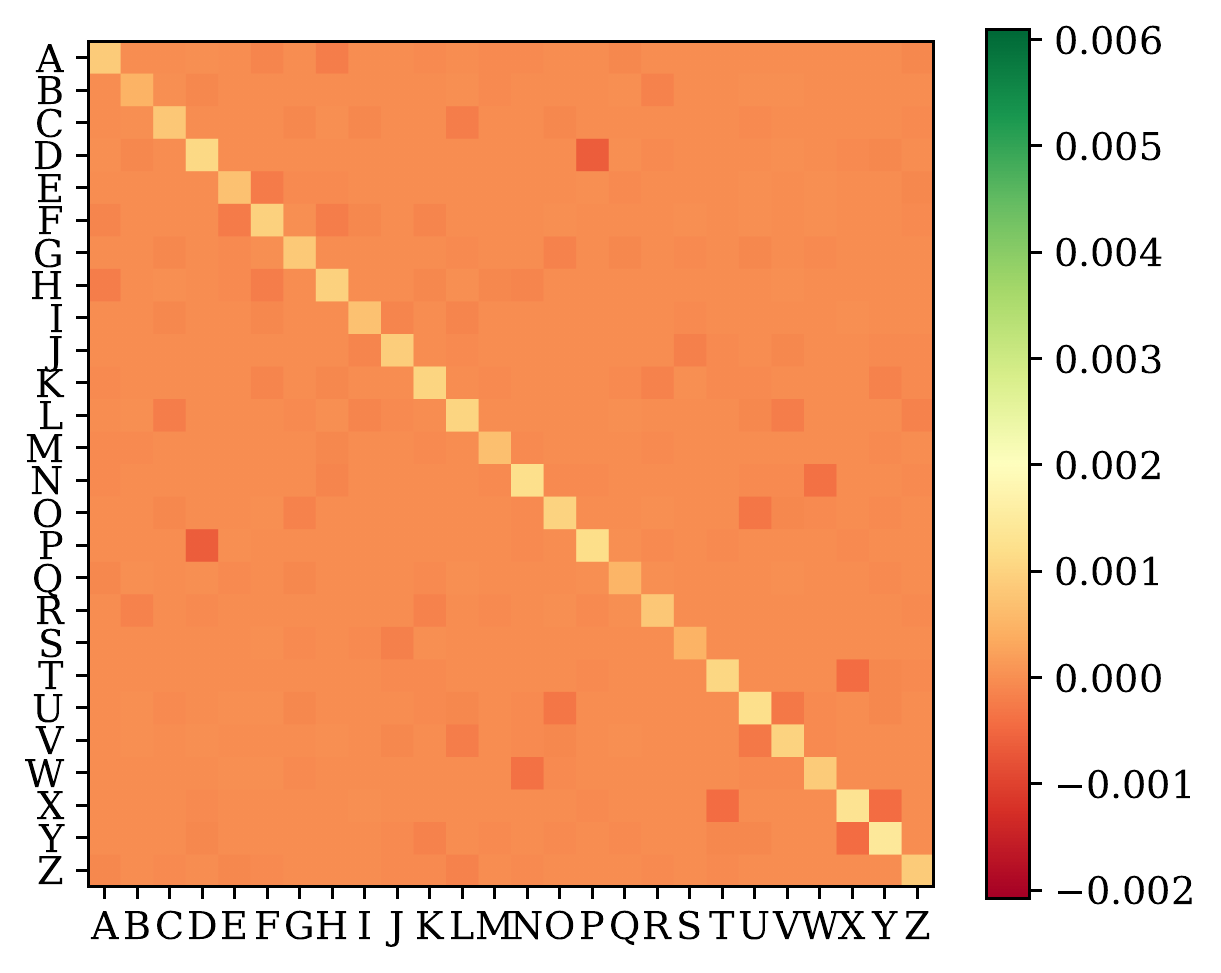} 
    \vspace{-0.4cm}
    \caption{Epistemic uncertainty.}
    \label{fig_app_epis_upp_app}
\end{subfigure}
\hfill
\begin{subfigure}{\subfigsizeb\textwidth}
    \centering
    \includegraphics[width=1\linewidth]{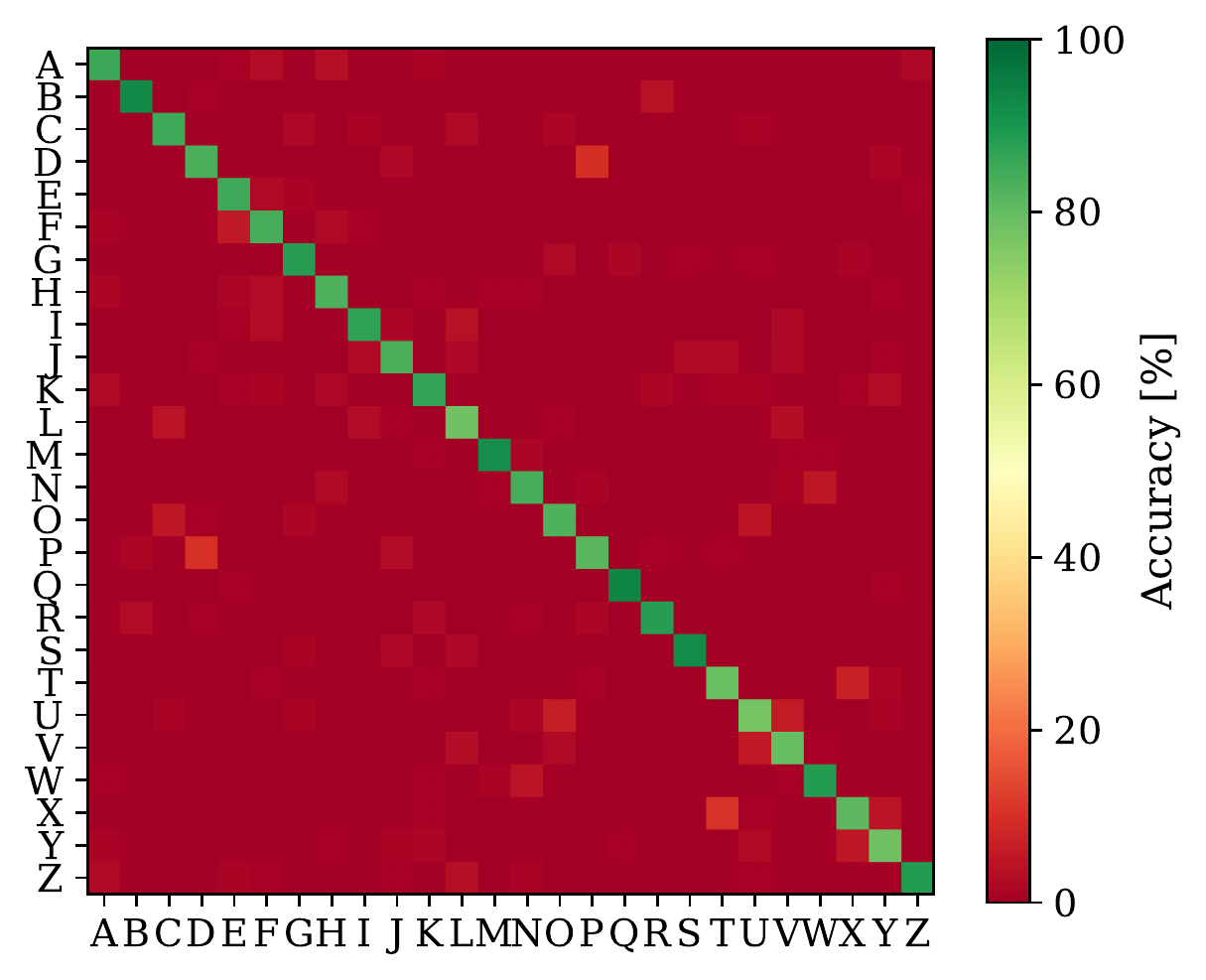} 
    \vspace{-0.4cm}
    \caption{Confusion matrix of accuracy.}
    \label{fig_app_conf_upp_app}
\end{subfigure}
\caption{Uncertainty prediction for the SWAG CNN+TCN model trained on the uppercase WD (right-handed only) dataset. Note that the color scale is fixed for all subplots for comparability with the other heatmaps.}
\label{fig:uppercase_conf_app}
\end{figure*}

\begin{figure*}[t!]
\centering
\begin{subfigure}{\subfigsizeb\textwidth}
    \centering
    \includegraphics[width=1\linewidth]{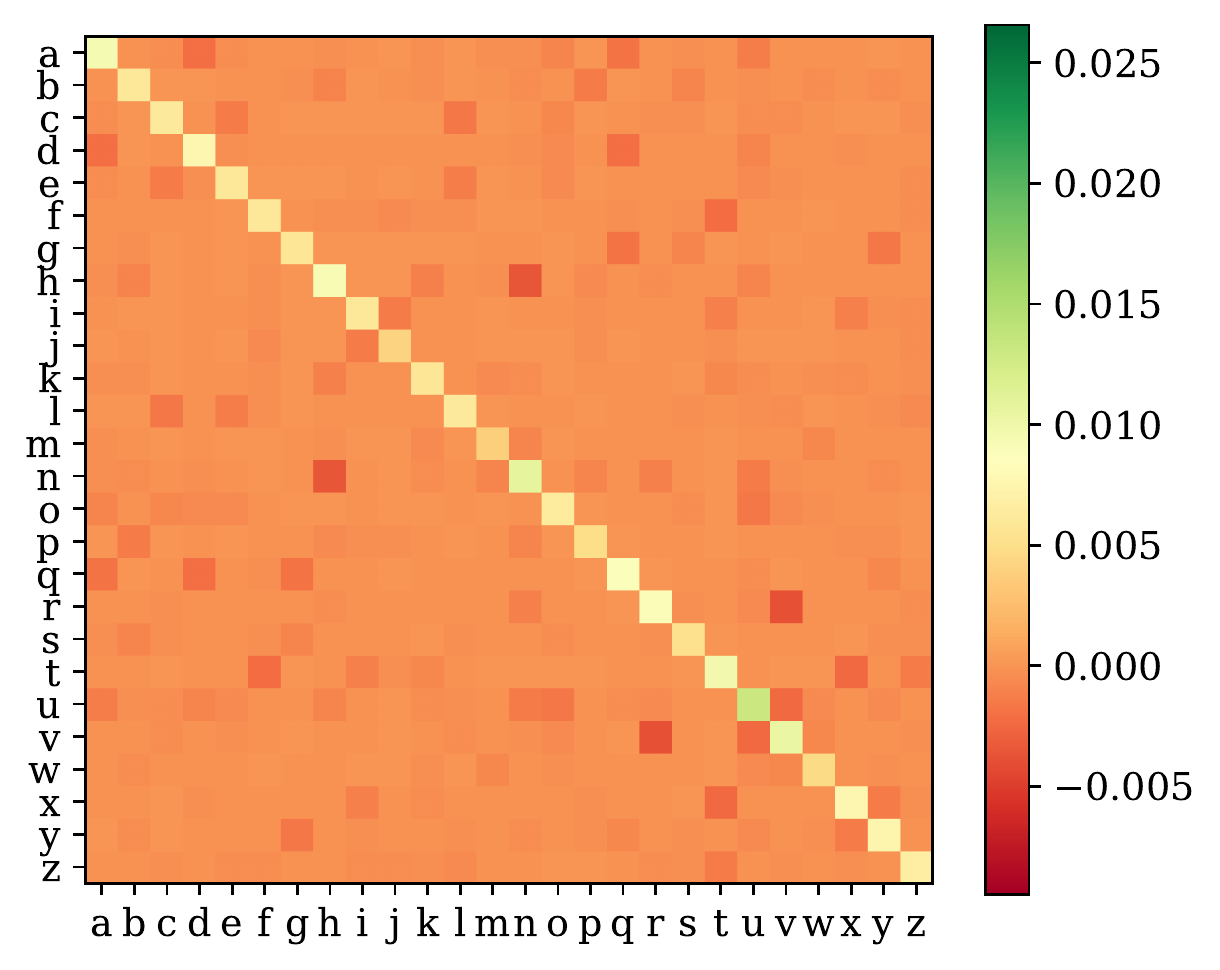}
    \vspace{-0.4cm}
    \caption{Aleatoric uncertainty.}
    \label{fig_app_alea_low_app}
\end{subfigure}
\hfill
\begin{subfigure}{\subfigsizeb\textwidth}
    \centering
    \includegraphics[width=1\linewidth]{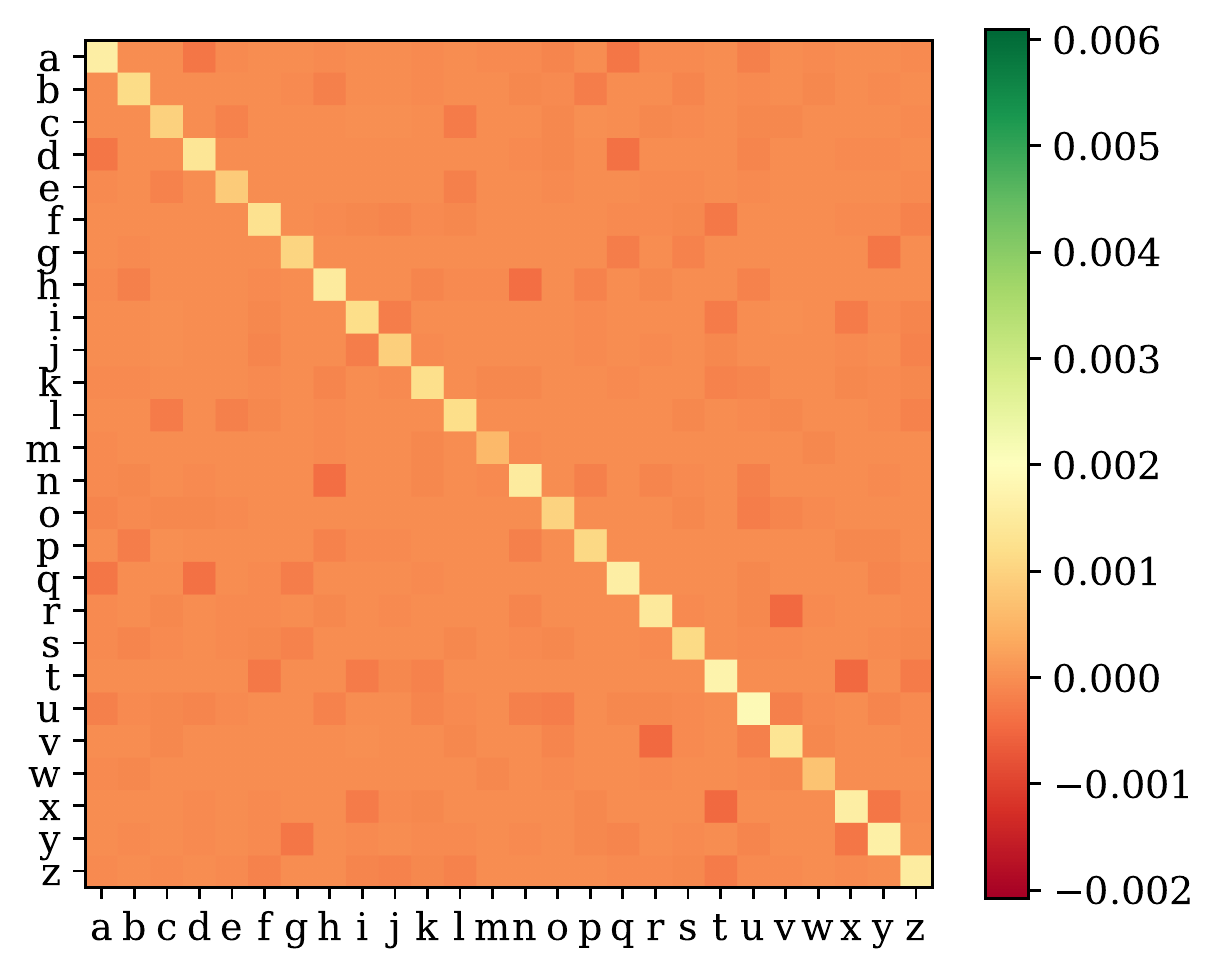} 
    \vspace{-0.4cm}
    \caption{Epistemic uncertainty.}
    \label{fig_app_epis_low_app}
\end{subfigure}
\hfill
\begin{subfigure}{\subfigsizeb\textwidth}
    \centering
    \includegraphics[width=1\linewidth]{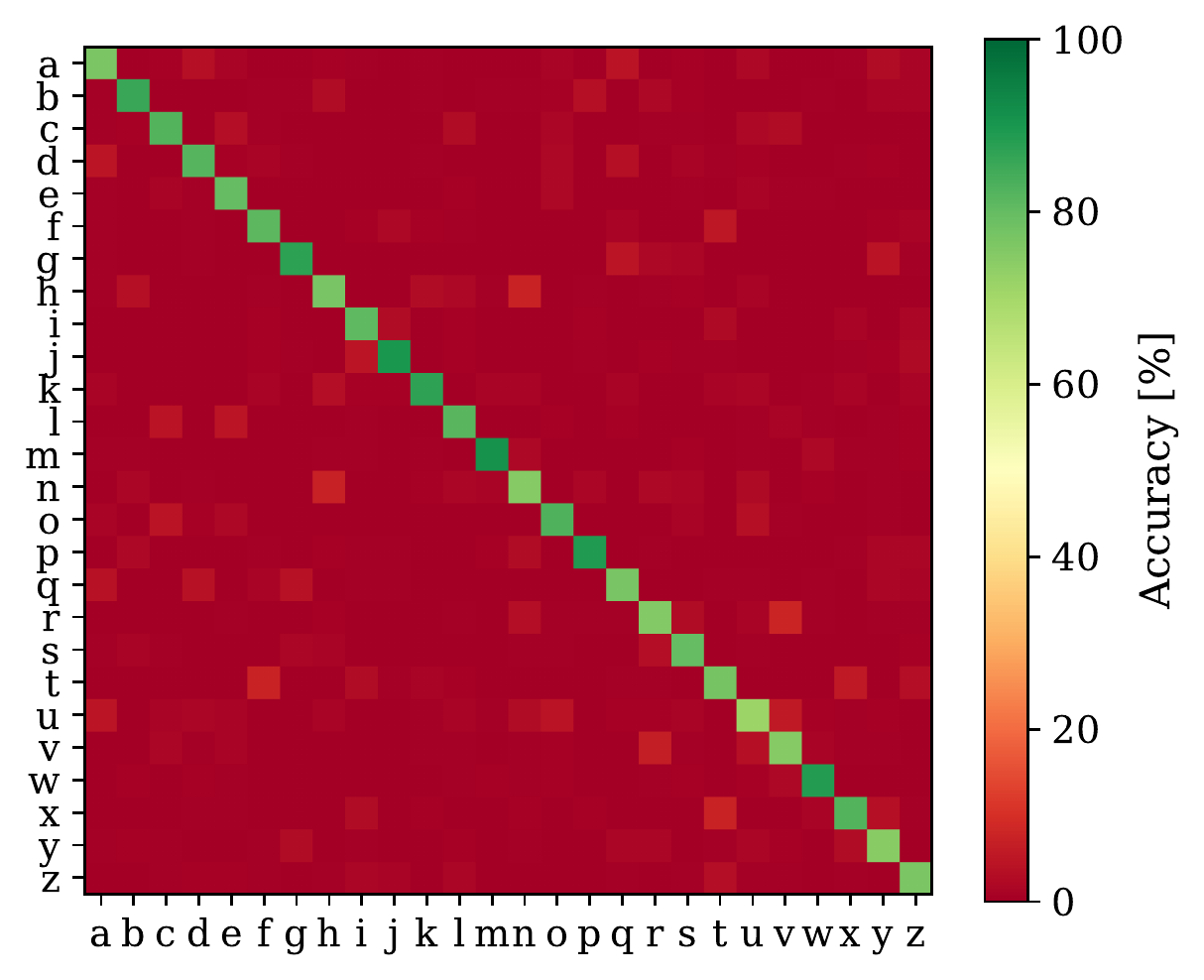} 
    \vspace{-0.4cm}
    \caption{Confusion matrix of accuracy.}
    \label{fig_app_conf_low_app}
\end{subfigure}
\caption{Uncertainty prediction for the SWAG CNN+TCN model trained on the lowercase WD (right-handed only) dataset. Note that the color scale is fixed for all subplots for comparability with the other heatmaps.}
\label{fig:lowercase_conf_app}
\end{figure*}

This section provides plots for the SWAG model that can directly be compared to the previously shown Deep Ensemble model plots. We observe very similar results between SWAG and Deep Ensemble models. Figure~\ref{plot_mi_entropy_swag} shows the MI and entropy for the SWAG model with the same pattern as for the Deep Ensemble model with lower absolute values. In Figure~\ref{fig:cal1_app}, we see the same overconfidence on left-handed data for SWAG models that have never seen this data similar as for Deep Ensemble models. The ECE by the SWAG model is marginally lower than the ECE by the Deep Ensemble model, but follows the same trend. The heatmaps in Figures~\ref{fig:both_app} for lowercase and uppercase characters, in Figure~\ref{fig:uppercase_conf_app} for uppercase characters only, and in Figure~\ref{fig:lowercase_conf_app} for lowercase characters only of the SWAG model show the same pattern as the heatmaps for Deep Ensemble models.

\end{document}